\documentclass[conference]{IEEEtran}
\IEEEoverridecommandlockouts
\usepackage{amsmath,amssymb,amsfonts}
\usepackage{algorithmic}
\usepackage{graphicx}
\usepackage{textcomp}
\usepackage{xcolor}
\def\BibTeX{{\rm B\kern-.05em{\sc i\kern-.025em b}\kern-.08em
    T\kern-.1667em\lower.7ex\hbox{E}\kern-.125emX}}

\usepackage{color}
\usepackage{fancyhdr}
\usepackage[hyphens]{url}
\usepackage{adjustbox}
\usepackage{hyperref}
\usepackage{enumitem}
\usepackage{xcolor}
\usepackage{lipsum}
\usepackage{multirow}
\usepackage{booktabs} 
\usepackage[bordercolor=white,backgroundcolor=gray!30,linecolor=black,colorinlistoftodos]{todonotes}
\usepackage{xspace}

\hypersetup{
     colorlinks = true,
     linkcolor = blue,
     anchorcolor = blue,
     citecolor = blue,
     filecolor = blue,
     urlcolor = blue
     }

\newcommand{\minihead}[1]{{\vspace{.5em}\noindent\textbf{#1}\xspace}}

\begin{document}

\title{MLPerf Inference Benchmark}

\author{\IEEEauthorblockN{
            Vijay Janapa Reddi,\IEEEauthorrefmark{1}
            Christine Cheng,\IEEEauthorrefmark{2}
            David Kanter,\IEEEauthorrefmark{3}
            Peter Mattson,\IEEEauthorrefmark{4}\\
            Guenther Schmuelling,\IEEEauthorrefmark{5}
            Carole-Jean Wu,\IEEEauthorrefmark{6}
            Brian Anderson,\IEEEauthorrefmark{4}
            Maximilien Breughe,\IEEEauthorrefmark{7}\\
            Mark Charlebois,\IEEEauthorrefmark{8}
            William Chou,\IEEEauthorrefmark{8}
            Ramesh Chukka,\IEEEauthorrefmark{2}
            Cody Coleman,\IEEEauthorrefmark{9}
            Sam Davis,\IEEEauthorrefmark{10}\\
            Pan Deng,\IEEEauthorrefmark{27}
            Greg Diamos,\IEEEauthorrefmark{11}
            Jared Duke,\IEEEauthorrefmark{4}
            Dave Fick,\IEEEauthorrefmark{12}
            J. Scott Gardner,\IEEEauthorrefmark{13}
            Itay Hubara,\IEEEauthorrefmark{14}\\
            Sachin Idgunji,\IEEEauthorrefmark{7}
            Thomas B. Jablin,\IEEEauthorrefmark{4}
            Jeff Jiao,\IEEEauthorrefmark{15}
            Tom St. John,\IEEEauthorrefmark{22}
            Pankaj Kanwar,\IEEEauthorrefmark{4}\\
            David Lee,\IEEEauthorrefmark{16}
            Jeffery Liao,\IEEEauthorrefmark{17}
            Anton Lokhmotov,\IEEEauthorrefmark{18}
            Francisco Massa,\IEEEauthorrefmark{6}
            Peng Meng,\IEEEauthorrefmark{27}\\
            Paulius Micikevicius,\IEEEauthorrefmark{7}
            Colin Osborne,\IEEEauthorrefmark{19}
            Gennady Pekhimenko,\IEEEauthorrefmark{20}
            Arun Tejusve Raghunath Rajan,\IEEEauthorrefmark{2}\\
            Dilip Sequeira,\IEEEauthorrefmark{7}
            Ashish Sirasao,\IEEEauthorrefmark{21}
            Fei Sun,\IEEEauthorrefmark{23}
            Hanlin Tang,\IEEEauthorrefmark{2}
            Michael Thomson,\IEEEauthorrefmark{24}\\
            Frank Wei,\IEEEauthorrefmark{25}
            Ephrem Wu,\IEEEauthorrefmark{21}
            Lingjie Xu,\IEEEauthorrefmark{28}
            Koichi Yamada,\IEEEauthorrefmark{2}
            Bing Yu,\IEEEauthorrefmark{16}\\
            George Yuan,\IEEEauthorrefmark{7}
            Aaron Zhong,\IEEEauthorrefmark{15}
            Peizhao Zhang,\IEEEauthorrefmark{6}
            Yuchen Zhou\IEEEauthorrefmark{26}}\\

    \IEEEauthorblockA{\IEEEauthorrefmark{1}Harvard University
                      \IEEEauthorrefmark{2}Intel
                      \IEEEauthorrefmark{3}Real World Insights
                      \IEEEauthorrefmark{4}Google
                      \IEEEauthorrefmark{5}Microsoft
                      \IEEEauthorrefmark{6}Facebook
                      \IEEEauthorrefmark{7}NVIDIA
                      \IEEEauthorrefmark{8}Qualcomm\\
                      \IEEEauthorrefmark{9}Stanford University
                      \IEEEauthorrefmark{10}Myrtle
                      \IEEEauthorrefmark{11}Landing AI
                      \IEEEauthorrefmark{12}Mythic
                      \IEEEauthorrefmark{13}Advantage Engineering
                      \IEEEauthorrefmark{14}Habana Labs\\
                      \IEEEauthorrefmark{15}Alibaba T-Head
                      \IEEEauthorrefmark{16}Facebook (formerly at MediaTek)
                      \IEEEauthorrefmark{17}OPPO (formerly at Synopsys)
                      \IEEEauthorrefmark{18}dividiti
                      \IEEEauthorrefmark{19}Arm\\
                      \IEEEauthorrefmark{20}University of Toronto \& Vector Institute
                      \IEEEauthorrefmark{21}Xilinx
                      \IEEEauthorrefmark{22}Tesla
                      \IEEEauthorrefmark{23}Alibaba (formerly at Facebook)\\
                      \IEEEauthorrefmark{24}Centaur Technology
                      \IEEEauthorrefmark{25}Alibaba Cloud
                      \IEEEauthorrefmark{26}General Motors
                      \IEEEauthorrefmark{27}Tencent
                      \IEEEauthorrefmark{28}Biren Research (formerly at Alibaba)}
}


\maketitle

\begin{abstract}
Machine-learning (ML) hardware and software system demand is burgeoning. Driven by ML applications, the number of different ML inference systems has exploded. Over 100 organizations are building ML inference chips, and the systems that incorporate existing models span at least three orders of magnitude in power consumption and five orders of magnitude in performance; they range from embedded devices to data-center solutions. Fueling the hardware are a dozen or more software frameworks and libraries. The myriad combinations of ML hardware and ML software make assessing ML-system performance in an architecture-neutral, representative, and reproducible manner challenging. There is a clear need for industry-wide standard ML benchmarking and evaluation criteria. MLPerf Inference answers that call. In this paper, we present our benchmarking method for evaluating ML inference systems. Driven by more than 30 organizations as well as more than 200 ML engineers and practitioners, MLPerf prescribes a set of rules and best practices to ensure comparability across systems with wildly differing architectures. The first call for submissions garnered more than 600 reproducible inference-performance measurements from 14 organizations, representing over 30 systems that showcase a wide range of capabilities. The submissions attest to the benchmark's flexibility and adaptability.
\end{abstract}


\section{Introduction}
\label{sec:introduction}

Machine learning (ML) powers a variety of applications from computer vision~(\cite{he2016deep, goodfellow2014generative, liu2016ssd, krizhevsky2012imagenet}) and natural-language processing~(\cite{vaswani2017attention, devlin2018bert}) to self-driving cars~(\cite{xu2018pointfusion, badrinarayanan2017segnet}) and autonomous robotics~\cite{levine2018learning}. Although ML-model training has been a development bottleneck and a considerable expense~\cite{amodei2018ai}, inference has become a critical workload. Models can serve as many as 200 trillion queries and perform over 6 billion translations a day~\cite{lee2019accelerating}. To address these growing computational demands, hardware, software, and system developers have focused on inference performance for a variety of use cases by designing optimized ML hardware and software. Estimates indicate that over 100 companies are targeting specialized inference chips~\cite{kanter2019supercomputing}. By comparison, only about 20 companies are targeting training chips~\cite{basicmi}. 


Each ML system takes a unique approach to inference, trading off latency, throughput, power, and model quality. The result is many possible combinations of ML tasks, models, data sets, frameworks, tool sets, libraries, architectures, and inference engines, making the task of evaluating inference performance nearly intractable. The spectrum of tasks is broad, including but not limited to image classification and localization, object detection and segmentation, machine translation, automatic speech recognition, text to speech, and recommendations. Even for a specific task, such as image classification, many ML models are viable. These models serve in a variety of scenarios from taking a single picture on a smartphone to continuously and concurrently detecting pedestrians through multiple cameras in an autonomous vehicle. Consequently, ML tasks have vastly different quality requirements and real-time processing demands. Even the implementations of the model's functions and operations can be highly framework specific, and they increase the complexity of the design task. To quantify these tradeoffs, the ML field needs a benchmark that is architecturally neutral, representative, and reproducible.

Both academic and industrial organizations have developed ML inference benchmarks. Examples of prior art include AIMatrix~\cite{aimatrix}, EEMBC MLMark~\cite{eembc}, and AIXPRT~\cite{aixprt} from industry, as well as AI Benchmark~\cite{ignatov2019ai}, TBD~\cite{zhu2018tbd}, Fathom~\cite{adolf2016fathom}, and DAWNBench~\cite{coleman2017dawnbench} from academia. Each one has made substantial contributions to ML benchmarking, but they were developed without industry-wide input from ML-system designers. As a result, there is no consensus on representative machine learning models, metrics, tasks, and rules. 

In this paper, we present MLPerf Inference, a standard ML inference benchmark suite with proper metrics and a benchmarking method (that complements MLPerf Training~\cite{mattson2019mlperf}) to fairly measure the inference performance of ML hardware, software, and services. Industry and academia jointly developed the benchmark suite and its methodology using  input from researchers and developers at more than 30 organizations. 

Over 200 ML engineers and practitioners contributed to the effort. MLPerf Inference is a consensus on the best benchmarking techniques, forged by experts in architecture, systems, and machine learning. We explain why designing the right ML benchmarking metrics, creating realistic ML inference scenarios, and standardizing the evaluation methods enables realistic performance optimization for inference quality. 


\textbf{\textcircled{\footnotesize{1}} We picked representative workloads for reproducibility and accessibility.} The ML ecosystem is rife with models. Comparing and contrasting ML-system performance across these models is nontrivial because they vary dramatically in model complexity and execution characteristics. In addition, a name such as {ResNet-50} fails to uniquely or portably describe a model. Consequently, quantifying system-performance improvements with an unstable baseline is difficult.


A major contribution of MLPerf is selection of representative models that permit reproducible measurements. Based on industry consensus, MLPerf Inference comprises models that are mature and have earned community support. Because the industry has studied them and can build efficient systems, benchmarking is accessible and provides a snapshot of ML-system technology. MLPerf models are also open source, making them a potential research tool. 

\textbf{\textcircled{\footnotesize{2}} We identified scenarios for realistic evaluation.} ML inference systems range from deeply embedded devices to smartphones to data centers. They have a variety of real-world applications and many figures of merit, each requiring multiple performance metrics. The right metrics, reflecting production use cases, allow not just MLPerf but also publications to show how a practical ML system would perform.

MLPerf Inference consists of four evaluation scenarios: {single-stream}, {multistream}, {server}, and {offline}. We arrived at them by surveying MLPerf's broad membership, which includes both customers and vendors. These scenarios represent many critical inference applications. We show that performance can vary drastically under these scenarios and their corresponding metrics. MLPerf Inference provides a way to simulate the realistic behavior of the inference system under test; such a feature is unique among AI benchmarks.

\textbf{\textcircled{\footnotesize{3}} We prescribe target qualities and tail-latency bounds in accordance with real-world use cases.} Quality and performance are intimately connected for all forms of ML. System architectures occasionally sacrifice model quality to reduce latency, reduce total cost of ownership (TCO), or increase throughput. The tradeoffs among accuracy, latency, and TCO are application specific. Trading 1\% model accuracy for 50\% lower TCO is prudent when identifying cat photos, but it is risky when detecting pedestrians for autonomous vehicles. 

To reflect this aspect of real deployments, MLPerf defines model-quality targets. We established per-model and scenario targets for inference latency and model quality. The latency bounds and target qualities are based on input gathered from ML-system end users and ML practitioners. As MLPerf improves these parameters in accordance with industry needs, the broader research community can track them to stay relevant.
    
\textbf{\textcircled{\footnotesize{4}} We set permissive rules that allow users to show both hardware and software capabilities.} The community has embraced a variety of languages and libraries, so MLPerf Inference is a semantic-level benchmark. We specify the task and the rules, but we leave implementation to submitters. 

Our benchmarks allow submitters to optimize the reference models, run them through their preferred software tool chain, and execute them on their hardware of choice. Thus, MLPerf Inference has two divisions: closed and open. Strict rules govern the closed division, which addresses the lack of a standard inference-benchmarking workflow. The open division, on the other hand, allows submitters to change the model and demonstrate different performance and quality targets. 
    
\textbf{\textcircled{\footnotesize{5}} We present an inference-benchmarking method that allows the models to change frequently while preserving the aforementioned contributions.} ML evolves quickly, so the  challenge for any benchmark is not performing the tasks, but implementing a method that can withstand rapid change.

MLPerf Inference focuses heavily on the benchmark's modular design to make adding new models and tasks less costly while preserving the usage scenarios, target qualities, and infrastructure. As we show in Section~\ref{sec:results_open_division}, our design has allowed users to add new models easily. We plan to extend the scope to include more areas and tasks. We are working to add new models (e.g., recommendation and time series), new scenarios (e.g., ``burst'' mode), better tools (e.g., a mobile application), and better metrics (e.g., timing preprocessing) to better reflect the performance of the whole ML pipeline.

\textbf{Impact.} MLPerf Inference (version 0.5) was put to the test in October 2019. We received over 600 submissions from 14 organizations, spanning various tasks, frameworks, and platforms. Audit tests automatically evaluated the submissions and cleared 595 of them as valid. The results show a four-orders-of-magnitude performance variation ranging from embedded devices and smartphones to data-center systems, demonstrating how the different metrics and inference scenarios are useful in more robustly accessing AI inference accelerators. 

\textbf{Staying up to date.} Because ML is still evolving, we established a process to regularly maintain and update MLPerf Inference. See \url{mlperf.org} for the latest benchmarks, rules etc.



\section{Inference-Benchmarking Challenges}
\label{sec:background}

A useful ML benchmark must overcome three critical challenges: the diversity of models, the variety of deployment scenarios, and the array of inference systems.

\subsection{Diversity of Models}
\label{sec:background:diversitymodels}

\begin{figure}[t!]
    \centering
    \vspace*{-2pt}
    \includegraphics[width=\linewidth]{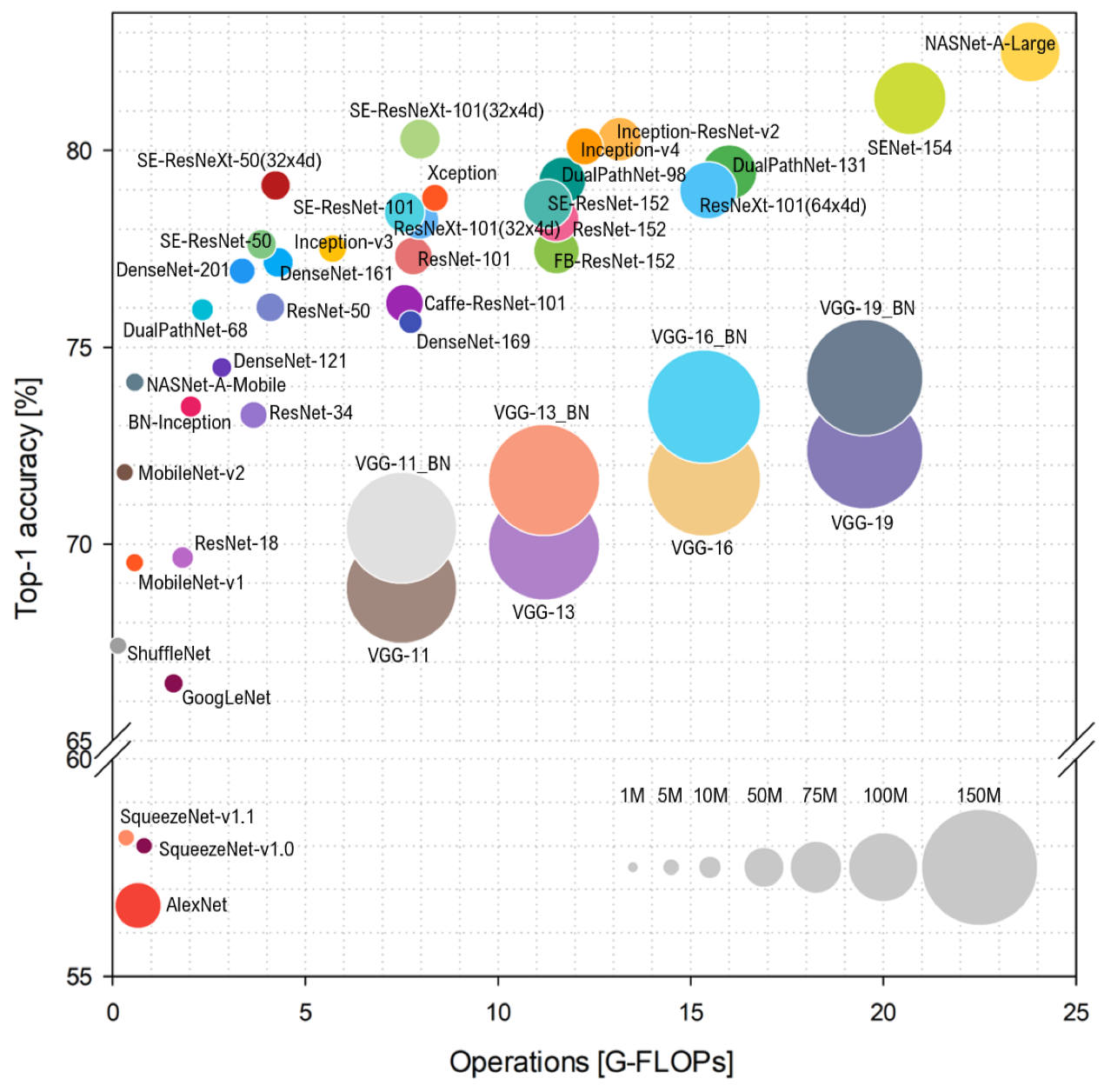}
    \caption{An example of ML-model diversity for image classification (plot from~\cite{bianco2018benchmark}). No single model is optimal; each one presents a design tradeoff between accuracy, memory requirements, and computational complexity.}
    \label{fig:ml_model}
    \vskip -0.15in
\end{figure}

Choosing the right models for a benchmark is difficult; the choice depends on the application. For example, pedestrian detection in autonomous vehicles has a much higher accuracy requirement than does labeling animals in photographs, owing to the different consequences of incorrect predictions. Similarly, quality-of-service (QoS) requirements for machine learning inference vary by several orders of magnitude from effectively no latency constraint for offline processes to milliseconds for real-time applications. 

Covering the vast design space necessitates careful selection of models that represent realistic scenarios. But even for a single ML task, such as image classification, numerous models present different tradeoffs between accuracy and computational complexity, as Figure \ref{fig:ml_model} shows. These models vary tremendously in compute and memory requirements (e.g., a 50$\times$ difference in gigaflops), while the corresponding Top-1 accuracy ranges from 55\% to 83\%~\cite{bianco2018benchmark}. This variation creates a Pareto frontier rather than one optimal choice. Even a small accuracy change (e.g., a few percent) can drastically alter the computational requirements (e.g., by 5--10$\times$). For example, SE-ResNeXt-50~(\cite{hu2018squeeze, xie2017aggregated}) and Xception~\cite{chollet2017xception} achieve roughly the same accuracy ($\sim$79\%) but exhibit a 2$\times$ computational difference.

\subsection{Deployment-Scenario Diversity}
\label{sec:background:diversityscenarios}

In addition to accuracy and computational complexity, a representative ML benchmark must take into account the input data's availability and arrival pattern across a variety of application-deployment scenarios. For example, in offline batch processing, such as photo categorization, all the data may be readily available in (network) storage, allowing accelerators to reach and maintain peak performance. By contrast, translation, image tagging, and other tasks experience variable arrival patterns based on end-user traffic. 

Similarly, real-time applications such as augmented reality and autonomous vehicles handle a constant flow of data rather than having it all in memory. Although the same model architecture could serve in each scenario, data batching and similar optimizations may be inapplicable, leading to drastically different performance. Timing the on-device inference latency alone fails to reflect the real-world requirements. 

\subsection{Inference-System Diversity}
\label{sec:background:diversitysystems}

\begin{figure}[t]
    \centering
    \includegraphics[trim=180 0 175 0, clip, width=\linewidth]{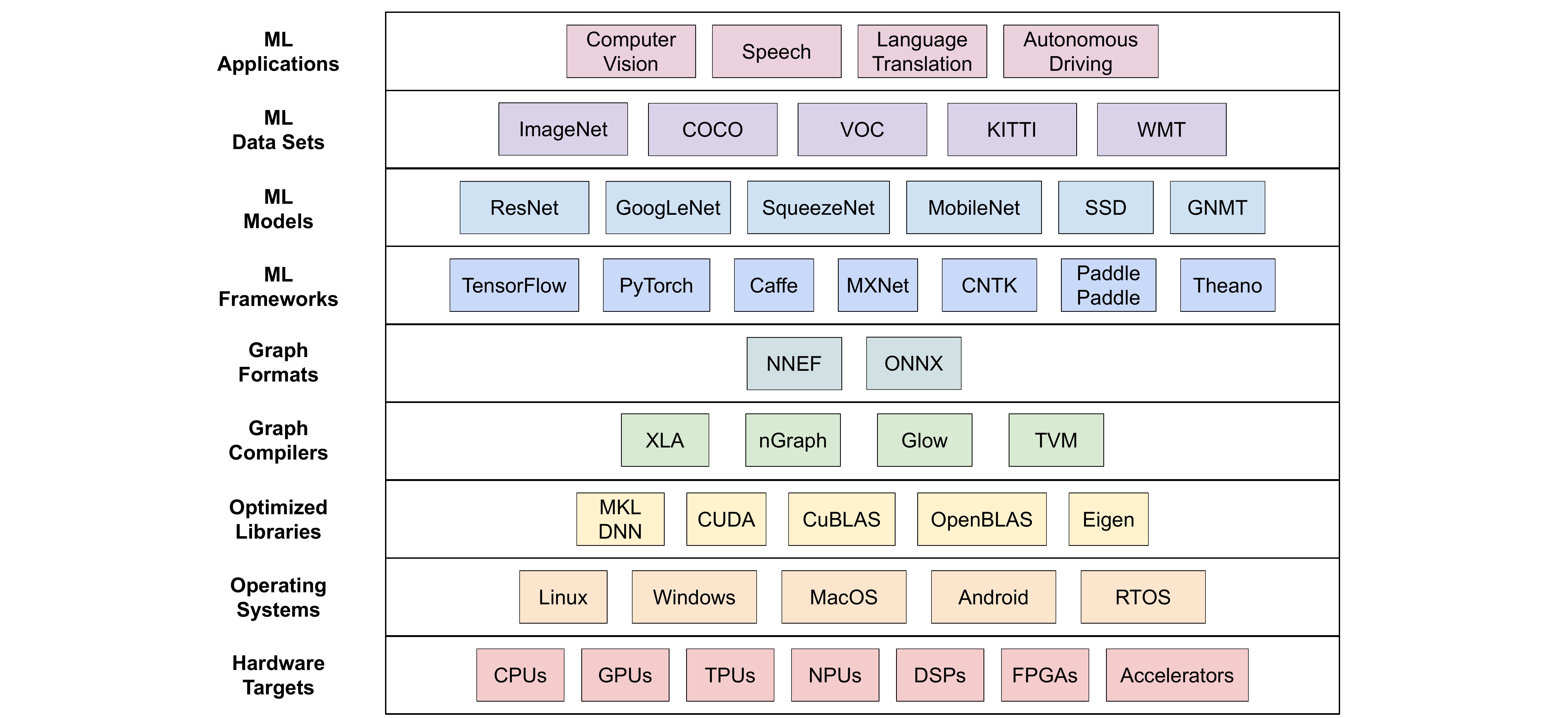}
    \caption{The diversity of options at every level of the stack, along with the combinations across the layers, make benchmarking inference systems hard.}
    \label{fig:hw_sw_stack}
    \vskip -0.15in
\end{figure}

The possible combinations of inference applications, data sets, models, machine-learning frameworks, tool sets, libraries, systems, and platforms are numerous, further complicating systematic and reproducible benchmarking. Figure~\ref{fig:hw_sw_stack} shows the wide breadth and depth of the ML space. The hardware and software side both exhibit substantial complexity.

\begin{table*}[t] 
 \caption{ML tasks in MLPerf Inference v0.5. Each one reflects critical commercial and research use cases for a large class of submitters, and together they cover a broad set of computing motifs (e.g., CNNs and RNNs).}
 \label{tab:ml_task}
  \vskip 0.05in
  \begin{center}
 \begin{small}
 \begin{sc}
 \begin{adjustbox}{width=\textwidth, center}
\begin{tabular}{c | c | c |c | c }
\toprule
\textbf{Area} & \textbf{Task} & \textbf{Reference Model} & \textbf{Data Set} & \textbf{Quality Target} \\
\midrule
Vision & Image classification (heavy) & \begin{tabular}[c]{@{}c@{}}ResNet-50 v1.5\\ ~25.6M parameters\\ 8.2 GOPS / input\end{tabular}
& ImageNet (224x224)
& 99\% of FP32 (76.456\%) Top-1 accuracy \\
\hline
Vision & Image classification (light) & \begin{tabular}[c]{@{}c@{}}MobileNet-v1 224\\ ~4.2M parameters\\ 1.138 GOPS / input\end{tabular}
& ImageNet (224x224)
& 98\% of FP32 (71.676\%) Top-1 accuracy \\
\hline
Vision & Object detection (heavy) & \begin{tabular}[c]{@{}c@{}}SSD-ResNet-34\\ 36.3M parameters\\ 433 GOPS / input\end{tabular}
& COCO (1,200x1,200)
& 99\% of FP32 (0.20 mAP) \\
\hline
Vision & Object detection (light) & \begin{tabular}[c]{@{}c@{}}SSD-MobileNet-v1\\ 6.91M parameters \\ 2.47 GOPS / input \end{tabular}
& COCO (300x300) 
& 99\% of FP32 (0.22 mAP) \\
\hline
Language & Machine translation & \begin{tabular}[c]{@{}c@{}}GNMT\\ ~210M parameters \end{tabular}
& WMT16 EN-DE
& 99\% of FP32 (23.9 SacreBleu)\\

\bottomrule
\end{tabular}
\end{adjustbox}
\end{sc}
\end{small}
\end{center}
\vskip -0.15in
\end{table*}

On the software side, about a dozen ML frameworks commonly serve for developing deep-learning models, such as Caffe/Caffe2~\cite{jia2014caffe}, Chainer~\cite{tokui2015chainer}, CNTK~\cite{seide2016cntk}, Keras~\cite{chollet2015keras}, MXNet~\cite{chen2015mxnet}, TensorFlow~\cite{abadi2016tensorflow}, and PyTorch~\cite{paszke2017pytorch}. Independently, there are also many optimized libraries, such as cuDNN~\cite{chetlur2018cudnn}, Intel~MKL~\cite{intel2018mkl}, and FBGEMM~\cite{khudia2018fbgemm}, supporting various inference run times, such as Apple~CoreML~\cite{apple2017coreml}, Intel OpenVino~\cite{intel2018openvino}, NVIDIA TensorRT~\cite{nvidiaYYYYtensorrt}, ONNX Runtime~\cite{bai2019onnx}, Qualcomm~SNPE~\cite{qualcommYYYYsnpe}, and TF-Lite~\cite{lee2019device}. 
Each combination has idiosyncrasies that make supporting the most current neural-network model architectures a challenge. Consider the non-maximum-suppression (NMS) operator for object detection. When training object-detection models in TensorFlow, the regular NMS operator smooths out imprecise bounding boxes for a single object. But this implementation is unavailable in TensorFlow Lite, which is tailored for mobile and instead implements fast NMS. As a result, when converting the model from TensorFlow to TensorFlow Lite, the accuracy of SSD-MobileNet-v1 decreases from 23.1\% to 22.3\% mAP. Such subtle differences make it hard to port models exactly from one framework to another. 

On the hardware side, platforms are tremendously diverse, ranging from familiar processors (e.g., CPUs, GPUs, and DSPs) to FPGAs, ASICs, and exotic accelerators such as analog and mixed-signal processors. Each platform comes with hardware-specific features and constraints that enable or disrupt performance depending on the model and scenario. 

%
%
\section{Benchmark Design}
\label{sec:goals}

Combining model diversity with the range of software systems presents a unique challenge to deriving a robust ML benchmark that meets industry needs. To overcome that challenge, we adopted a set of principles for developing a robust yet flexible offering based on community input. In this section, we describe the benchmarks, the quality targets, and the scenarios under which the ML systems can be evaluated.

\subsection{Representative, Broadly Accessible Workloads}

Designing ML benchmarks is different from designing traditional non-ML benchmarks. MLPerf defines high-level tasks (e.g., image classification) that a machine-learning system can perform. For each we provide one or more canonical reference models in a few widely used frameworks. Any implementation that is mathematically equivalent to the reference model is considered valid, and certain other deviations (e.g., numerical formats) are also allowed. For example, a fully connected layer can be implemented with different cache-blocking and evaluation strategies. Consequently, submitted results require optimizations to achieve good performance.

A reference model and a valid class of equivalent implementations gives most ML systems freedom while still enabling relevant comparisons. MLPerf provides reference models using 32-bit floating-point weights and, for convenience, carefully implemented equivalent models to address three formats: TensorFlow~\cite{abadi2016tensorflow}, PyTorch~\cite{paszke2017pytorch}, and ONNX~\cite{bai2019onnx}.


As Table~\ref{tab:ml_task} illustrates, we chose an initial set of vision and language tasks along with associated reference models. Together, vision and translation serve widely across computing systems, from edge devices to cloud data centers. Mature and well-behaved reference models with different architectures (e.g., CNNs and RNNs) were available, too. 




\textbf{Image classification.} Many commercial applications employ image classification, which is a de facto standard for evaluating ML-system performance. A classifier network takes an image and selects the class that best describes it. Example applications include photo searches, text extraction, and industrial automation, such as object sorting and defect detection. We use the ImageNet 2012 data set~\cite{deng2009imagenet}, crop the images to 224x224 in preprocessing, and measure Top-1 accuracy.

\begin{table*}[t] 
 \caption{Scenario description and metrics. Each scenario targets a real-world use case based on customer and vendor input.}
 \label{tab:scenarios}
 \vskip 0.05in
  \begin{center}
 \begin{sc}
 \begin{adjustbox}{width=\textwidth, center}
\begin{tabular}{c | c | c |c | c }
\toprule
\textbf{Scenario} & \textbf{Query Generation} & \textbf{Metric} & \textbf{Samples/Query} & \textbf{Examples} \\
\midrule
Single-stream (SS) & Sequential
& 90th-percentile latency
& 1 & \begin{tabular}[c]{@{}c@{}}Typing autocomplete,\\ real-time AR\end{tabular}  \\
\hline
Multistream (MS) & Arrival interval with dropping
& \begin{tabular}[c]{@{}c@{}}Number of streams\\ subject to latency bound\end{tabular} 
& $N$ & \begin{tabular}[c]{@{}c@{}}Multicamera driver assistance,\\ large-scale automation\end{tabular}\\
\hline
Server (S) & Poisson distribution
& \begin{tabular}[c]{@{}c@{}}Queries per second\\ subject to latency bound\end{tabular} 
& 1 & Translation website \\
\hline
Offline (O) & Batch
& Throughput 
& At least 24,576 & Photo categorization \\

\bottomrule
\end{tabular}
\end{adjustbox}
\end{sc}
\end{center}
\vskip -0.15in
\end{table*}

We selected two models: a computationally heavyweight model that is more accurate and a computationally lightweight model that is faster but less accurate. The heavyweight model, ResNet-50 v1.5~(\cite{he2016deep, mlperf2017resnet}), comes directly from the MLPerf Training suite to maintain alignment. ResNet-50 is the most common network for performance claims. Unfortunately, it has multiple subtly different implementations that make most comparisons difficult. We specifically selected ResNet-50 v1.5 to ensure useful comparisons and compatibility across major frameworks. This network exhibits good reproducibility, making it a low-risk choice.

The lightweight model, MobileNet-v1-224~\cite{howard2017mobilenets}, employs smaller, depth-wise-separable convolutions to reduce the complexity and computational burden. MobileNet networks offer varying compute and accuracy options---we selected the full-width, full-resolution MobileNet-v1-1.0-224. It reduces the parameters by 6.1$\times$ and the operations by 6.8$\times$ compared with ResNet-50 v1.5. We evaluated both MobileNet-v1 and MobileNet-v2~\cite{sandler2018mobilenetv2} for the MLPerf Inference v0.5 suite, selecting the former because of its wider adoption.

\textbf{Object detection.} Object detection is a vision task that determines the coordinates of bounding boxes around objects in an image and then classifies those boxes. Implementations typically use a pretrained image-classifier network as a backbone or feature extractor, then perform regression for localization and bounding-box selection. Object detection is crucial for automotive tasks, such as detecting hazards and analyzing traffic, and for mobile-retail tasks, such as identifying items in a picture. We chose the COCO data set~\cite{lin2014microsoft} with both a lightweight model and a heavyweight model.

Similar to image classification, we selected two models. Our small model uses the 300x300 image size, which is typical of resolutions in smartphones and other compact devices. For the larger model, we upscale the data set to more closely represent the output of a high-definition image sensor (1.44~MP total). The choice of the larger input size is based on community feedback, especially from automotive and industrial-automation customers. The quality metric for object detection is mean average precision (mAP). 

The heavyweight object detector's reference model is SSD~\cite{liu2016ssd} with a ResNet-34 backbone, which also comes from our training benchmark. The lightweight object detector's reference model uses a MobileNet-v1-1.0 backbone, which is more typical for constrained computing environments. We selected the MobileNet feature detector on the basis of feedback from the mobile and embedded communities.

\textbf{Translation.} Neural machine translation (NMT) is popular in natural-language processing. NMT models translate a sequence of words from a source language to a target language and appear in translation applications and services. Our translation data set is WMT16 EN-DE~\cite{wmt2016}. The quality measurement is the Bilingual Evaluation Understudy (BLEU) score~\cite{papineni2002bleu}, implemented using  SacreBLEU~\cite{post2018call}. Our reference model is GNMT~\cite{wu2016google}, which employs a well-established recurrent-neural-network (RNN) architecture and is part of the training benchmark. RNNs are popular for sequential and time-series data, so including GNMT ensures our reference suite captures a variety of compute motifs.

\subsection{Robust Quality Targets}

Quality and performance are intimately connected for all forms of machine learning. Although the starting point for inference is a pretrained reference model that achieves a target quality, many system architectures can sacrifice model quality to reduce latency, reduce total cost of ownership (TCO), or increase throughput. The tradeoffs between accuracy, latency, and TCO are application specific. Trading 1\% model accuracy for 50\% lower TCO is prudent when identifying cat photos, but it is less so during online pedestrian detection. To reflect this important aspect, we established per-model quality targets.

We require that almost all implementations achieve a quality target within 1\% of the FP32 reference model's accuracy. (For example, the ResNet-50 v1.5 model achieves 76.46\% Top-1 accuracy, and an equivalent model must achieve at least 75.70\% Top-1 accuracy.) Initial experiments, however, showed that for mobile-focused networks---MobileNet and SSD-MobileNet---the accuracy loss was unacceptable without retraining. We were unable to proceed with the low accuracy as performance benchmarking would become unrepresentative.

To address the accuracy drop, we took three steps. First, we trained the MobileNet models for quantization-friendly weights, enabling us to narrow the quality window to 2\%. Second, to reduce the training sensitivity of MobileNet-based submissions, we provided equivalent MobileNet and SSD-MobileNet implementations quantized to an 8-bit integer format. Third, for SSD-MobileNet, we reduced the quality requirement to 22.0~mAP to account for the challenges of using a MobileNet backbone.

To improve the submission comparability, we disallow retraining. Our prior experience and feasibility studies confirmed that for 8-bit integer arithmetic, which was an expected deployment path for many systems, the $\sim$1\% relative-accuracy target was easily achievable without retraining. 

\subsection{Realistic End-User Scenarios}

ML applications have a variety of usage models and many figures of merit, which in turn require multiple performance metrics. For example, the figure of merit for an image-recognition system that classifies a video camera's output will be entirely different than for a cloud-based translation system. 

To address these scenarios, we surveyed MLPerf's broad membership, which includes both customers and vendors. On the basis of that feedback, we identified four scenarios that represent many critical inference applications: {single-stream}, {multistream}, {server}, and {offline}. These scenarios emulate the ML-workload behavior of mobile devices, autonomous vehicles, robotics, and cloud-based setups (Table~\ref{tab:scenarios}). 

\textbf{Single-stream.} The single-stream scenario represents one inference-query stream with a query sample size of 1, reflecting the many client applications where responsiveness is critical. An example is offline voice transcription on Google's Pixel~4 smartphone. To measure performance, we inject a single query into the inference system; when the query is complete, we record the completion time and inject the next query. The metric is the query stream's 90th-percentile latency. 

\textbf{Multistream.} The multistream scenario represents applications with a stream of queries, but each query comprises multiple inferences, reflecting a variety of industrial-automation and remote-sensing tasks. For example, many autonomous vehicles analyze frames from multiple cameras simultaneously. To model a concurrent scenario, we send a new query comprising \emph{N} input samples at a fixed time interval (e.g., 50 ms). The interval is benchmark specific and also acts as a latency bound that ranges from 50 to 100 milliseconds. If the system is available, it processes the incoming query. If it is still processing the prior query in an interval, it skips that interval and delays the remaining queries by one interval. No more than 1\% of the queries may produce one or more skipped intervals. A query's \emph{N} input samples are contiguous in memory, which accurately reflects production input pipelines and avoids penalizing systems that would otherwise require copying of samples to a contiguous memory region before starting inference. The performance metric is the integer number of streams that the system supports while meeting the QoS requirement.

\textbf{Server.} The server scenario represents online applications where query arrival is random and latency is important. Almost every consumer-facing website is a good example, including services such as online translation from Baidu, Google, and Microsoft. For this scenario, queries have one sample each, in accordance with a Poisson distribution. The system under test responds to each query within a benchmark-specific latency bound that varies from 15 to 250 milliseconds. No more than 1\% of queries may exceed the latency bound for the vision tasks and no more than 3\% may do so for translation. The server scenario's performance metric is the Poisson parameter that indicates the queries-per-second (QPS) achievable while meeting the QoS requirement.

\begin{table}[t!] 
 \caption{Latency constraints in the multistream and server scenarios.}
 \label{tab:latency}
 \vskip 0.05in
  \begin{center}
 \begin{small}
 \begin{sc}
 \begin{adjustbox}{width=\columnwidth, center}
\begin{tabular}{c | c | c  }
\toprule
\textbf{Task} & \begin{tabular}[c]{@{}c@{}}\textbf{Multistream}\\ \textbf{Arrival Time}\end{tabular} & \begin{tabular}[c]{@{}c@{}}\textbf{Server QoS}\\ \textbf{Constraint}\end{tabular}\\
\midrule
Image classification (heavy) & ~~50 ms & ~~15 ms \\
\hline
Image classification (light) & ~~50 ms & ~~10 ms \\
\hline
Object detection (heavy) & ~~66 ms & 100 ms \\
\hline
Object detection (light) & ~~50 ms & ~~10 ms \\
\hline
Machine translation & 100 ms & 250 ms \\

\bottomrule
\end{tabular}
\end{adjustbox}
\end{sc}
\end{small}
\end{center}
\vskip -0.15in
\end{table}

\textbf{Offline.} The offline scenario represents batch-processing applications where all data is immediately available and latency is unconstrained. An example is identifying the people and locations in a photo album. For this scenario, we send a single query that includes all sample-data IDs to be processed, and the system is free to process the input data in any order. Similar to the multistream scenario, neighboring samples in the query are contiguous in memory. The metric for the offline scenario is throughput measured in samples per second.

For the multistream and server scenarios, latency is a critical component of the system behavior and constrains various performance optimizations. For example, most inference systems require a minimum (and architecture-specific) batch size to fully utilize the underlying computational resources. But the query arrival rate in servers is random, so they must optimize for tail latency and potentially process inferences with a suboptimal batch size. 

Table~\ref{tab:latency} shows the latency constraints for each task in MLPerf Inference v0.5. As with other aspects of the benchmark, we selected these constraints on the basis of feasibility and community consultation. The multistream scenario's arrival times for most vision tasks correspond to a frame rate of 15--20 Hz, which is a minimum for many applications. The server scenario's QoS constraints derive from estimates of the inference timing budget given an overall user latency target.

\subsection{Statistically Confident Tail-Latency Bounds}

Each task and scenario combination requires a minimum number of queries to ensure results are statistically robust and adequately capture steady-state system behavior. That number is determined by the tail-latency percentile, the desired margin, and the desired confidence interval. Confidence is the probability that a latency bound is within a particular margin of the reported result. We chose a 99\% confidence bound and set the margin to a value much less than the difference between the tail-latency percentage and 100\%. Conceptually, that margin ought to be relatively small. Thus, our selection is one-twentieth of the difference between the tail-latency percentage and 100\%. The equation is as follows:

\begin{small}
\begin{equation}
    Margin = \frac{1 - TailLatency}{20} 
\end{equation}  
\begin{equation}
\begin{aligned}
    {NumQueries} &= ({Normslnv}(\frac{1-{Confidence}}{2}))^{2}\\ &\times \frac{{TailLatency}\times({1 - TailLatency})}{{Margin}^2} 
\end{aligned}
\end{equation}
\end{small}

Equation 2 provides the number of queries that are necessary to achieve a statistically valid measurement. The math for determining the appropriate sample size for a latency-bound throughput experiment is the same as that for determining the appropriate sample size for an electoral poll given an infinite electorate where three variables determine the sample size: tail-latency percentage, confidence, and margin~\cite{tamhane2000statistics}.

Table \ref{tab:statsconf} shows the query requirements. The total query count and tail-latency percentile are scenario and task specific. The single-stream scenario requires 1,024 queries, and the offline scenario requires 1 query with at least 24,576 samples. The former has the fewest queries to execute, as we wanted the run time to be short enough that embedded platforms and smartphones could complete the benchmarks quickly. 

\begin{table}[t] 
 \caption{Query requirements for statistical confidence. All results must meet the minimum LoadGen scenario requirements.}
 \label{tab:statsconf}
    \vskip 0.05in
  \begin{center}
 \begin{small}
 \begin{sc}
 \begin{adjustbox}{width=\columnwidth, center}
\begin{tabular}{c | c | c | c | c}
\toprule
\begin{tabular}[c]{@{}c@{}}\textbf{Tail-Latency}\\ \textbf{Percentile}\end{tabular} & \begin{tabular}[c]{@{}c@{}}\textbf{Confidence}\\ \textbf{Interval}\end{tabular} & \begin{tabular}[c]{@{}c@{}}\textbf{Error}\\ \textbf{Margin}\end{tabular} & \textbf{Inferences} & \begin{tabular}[c]{@{}c@{}}\textbf{Rounded}\\ \textbf{Inferences}\end{tabular}\\
\midrule
90\% & 99\% & 0.50\% & 23,886 & $3\times2^{13} = 24,576$ \\
\hline
95\% & 99\% & 0.25\% & 50,425 & $7\times2^{13} = 57,344$\\
\hline
99\% & 99\% & 0.05\% & 262,742 & $33\times2^{13} = 270,336$ \\

\bottomrule
\end{tabular}
\end{adjustbox}
\end{sc}
\end{small}
\end{center}
\vskip -0.15in
\end{table} 

For scenarios with latency constraints, our goal is to ensure a 99\% confidence interval that the constraints hold. As a result, the benchmarks with more-stringent latency constraints require more queries in a highly nonlinear fashion. The number of queries is based on the aforementioned statistics and is rounded up to the nearest multiple of $2^{13}$. 

A 99th-percentile guarantee requires 262,742 queries, which rounds up to $33 \times 2^{13}$, or 270K. For both multistream and server, this guarantee for vision tasks requires 270K queries, as Table \ref{tab:task_queries} shows. Because a multistream benchmark will process $N$ samples per query, the total number of samples will be $N \times$ 270K. Machine translation has a 97th-percentile latency guarantee and requires only 90K queries.

For repeatability, we run the multistream and server scenarios several times. But the multistream scenario's arrival rate and query count  guarantee a 2.5- to 7.0-hour run time. To strike a balance between repeatability and run time, we require five runs for the server scenario, with the result being the minimum of these five. The other scenarios require one run. We expect to revisit this choice in future MLPerf versions.

All benchmarks must also run for at least 60 seconds and process additional queries and/or samples as the scenarios require. The minimum run time ensures we measure the equilibrium behavior of power-management systems and systems that support dynamic voltage and frequency scaling (DVFS), particularly for the single-stream scenario with few queries.

\section{Inference Submission System}
\label{sec:benchmarks}

\begin{figure}[t!]
    \centering
    \includegraphics[width=0.65\linewidth]{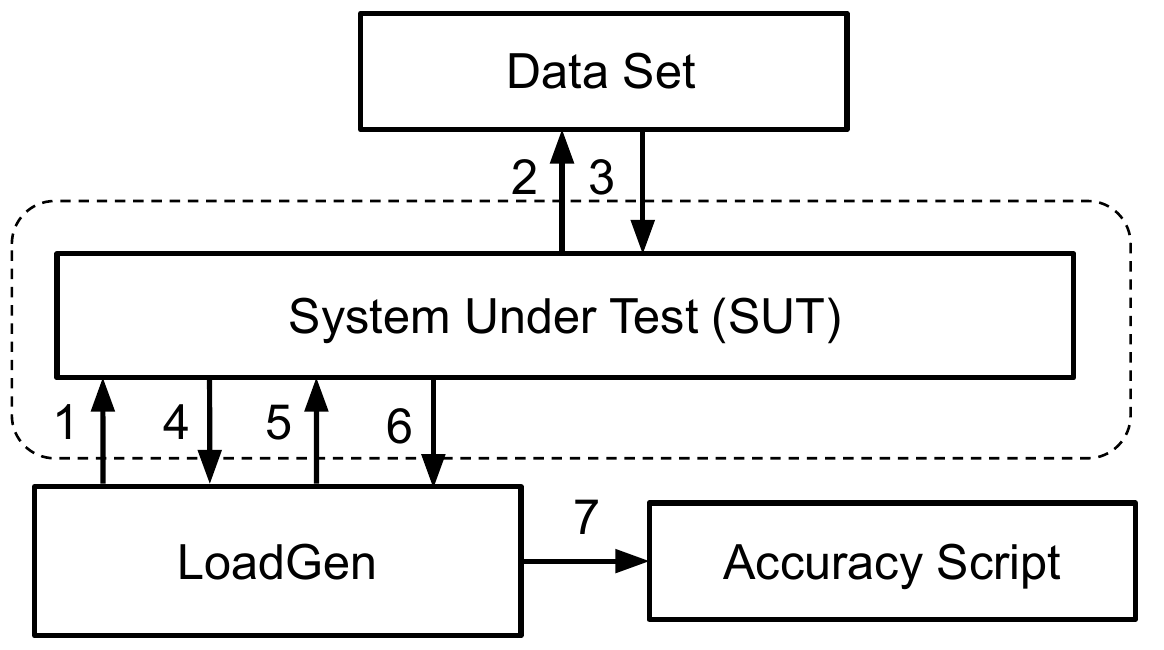}
    \caption{MLPerf Inference system under test (SUT) and associated components. First, the LoadGen requests that the SUT load samples (1). The SUT then loads samples into memory (2--3) and signals the LoadGen when it is ready (4). Next, the LoadGen issues requests to the SUT (5). The benchmark processes the results and returns them to the LoadGen (6), which then outputs logs for the accuracy script to read and verify (7).}
    \label{fig:SUT}
    \vskip -0.15in
\end{figure}

An MLPerf Inference submission system contains a system under test (SUT), the Load Generator (LoadGen), a data set, and an accuracy script. In this section we describe these various components. Figure~\ref{fig:SUT} shows an overview of an inference system. The data set, LoadGen, and accuracy script are fixed for all submissions and are provided by MLPerf. Submitters have wide discretion to implement an SUT according to their architecture's requirements and their engineering judgment. By establishing a clear boundary between submitter-owned and MLPerf-owned components, the benchmark maintains comparability among submissions.  

\subsection{System Under Test}

The submitter is responsible for the system under test. The goal of MLPerf Inference is to measure system performance across a wide variety of architectures. But realism, comparability, architecture neutrality, and friendliness to small submission teams require careful tradeoffs. For instance, some deployments have teams of compiler, computer-architecture, and machine-learning experts aggressively co-optimizing the training and inference systems to achieve cost, accuracy, and latency targets across a massive global customer base. An unconstrained benchmark would disadvantage companies with less experience and fewer ML-training resources.

Therefore, we set the model-equivalence rules to allow submitters to, for efficiency, reimplement models on different architectures. The rules provide a complete list of disallowed techniques and a list of allowed-technique examples. We chose an explicit blacklist to encourage a wide range of techniques and to support architectural diversity. The list of examples illustrates the blacklist boundaries while also encouraging common and appropriate optimizations. 

\begin{table}[t] 
 \caption{Number of queries and samples per query for each task.}
 \label{tab:task_queries}
   \vskip 0.05in
  \begin{center}
 \begin{sc}
 \begin{adjustbox}{width=\columnwidth, center}
\begin{tabular}{c | c | c | c | c}
\toprule
\multicolumn{1}{c|}{\multirow{2}{*}{{\textbf{Model}}}} & \multicolumn{4}{c}{\textbf{Number of Queries / Samples per Query}}   \\ \cline{2-5} 
\multicolumn{1}{c|}{}                   & \multicolumn{1}{l|}{\textbf{Single-Stream}} & \multicolumn{1}{l|}{\textbf{Multistream}} & \multicolumn{1}{l|}{\textbf{Server}} & \multicolumn{1}{l}{\textbf{Offline}} \\ 
\midrule
Image classification (heavy) & 1K / 1 & 270K / $N$ & 270K / 1 & 1 / 24K \\
\hline
Image classification (light) & 1K / 1 & 270K / $N$ & 270K / 1 & 1 / 24K \\
\hline
Object detection (heavy) & 1K / 1 & 270K / $N$ & 270K / 1 & 1 / 24K \\
\hline
Object detection (light) & 1K / 1 & 270K / $N$ & 270K / 1 & 1 / 24K \\
\hline
Machine translation & 1K / 1 & 90K / $N$ & 90K / 1 & 1 / 24K \\

\bottomrule
\end{tabular}
\end{adjustbox}
\end{sc}
\end{center}
\vskip -0.15in
\end{table}

\textbf{Allowed techniques.}
Examples of allowed techniques include arbitrary data arrangement as well as different input and in-memory representations of weights, mathematically equivalent transformations, approximations (e.g., replacing a transcendental function with a polynomial), out-of-order query processing within the scenario's limits, replacing dense operations with mathematically equivalent sparse operations, fusing and unfusing operations, and dynamically switching between one or more batch sizes.

To promote architecture and application neutrality, we adopted untimed preprocessing. Implementations may transform their inputs into system-specific ideal forms as an untimed operation. Ideally, a whole-system benchmark should capture all performance-relevant operations. In MLPerf, however, we explicitly allow untimed preprocessing. There is no vendor- or application-neutral preprocessing. For example, systems with integrated cameras can use hardware/software co-design to ensure that images reach memory in an ideal format; systems accepting JPEGs from the Internet cannot.

We also allow and enable quantization to many different numerical formats to ensure architecture neutrality. Submitters register their numerics ahead of time to help guide accuracy-target discussions. The approved list includes INT4, INT8, INT16, UINT8, UINT16, FP11 (1-bit sign, 5-bit mantissa, and 5-bit exponent), FP16, bfloat16, and FP32. Quantization to lower-precision formats typically requires calibration to ensure sufficient inference quality. For each reference model, MLPerf provides a small, fixed data set that can be used to calibrate a quantized network. Additionally, it offers MobileNet versions that are prequantized to INT8, since without retraining (which we disallow) the accuracy falls dramatically.

\textbf{Prohibited techniques.}
We prohibit retraining and pruning to ensure comparability. Although this restriction may fail to reflect realistic deployment in some cases, the interlocking requirements to use reference weights (possibly with calibration) and minimum accuracy targets are most important for comparability. We may eventually relax this restriction.

To simplify the benchmark evaluation, we disallow caching. In practice, inference systems cache queries. For example, ``I love you'' is one of Google Translate's most frequent queries, but the service does not translate the phrase ab initio each time. Realistically modeling caching in a benchmark, however, is difficult because cache-hit rates vary substantially with the application. Furthermore, our data sets are relatively small, and large systems could easily cache them in their entirety.  

We also prohibit optimizations that are benchmark or data-set aware and that are inapplicable to production environments. For example, real query traffic is unpredictable, but for the benchmark, the traffic pattern is predetermined by the pseudorandom-number-generator seed. Optimizations that take advantage of a fixed number of queries or that take the LoadGen implementation into account are prohibited. Similarly, any optimization employing the statistics of the performance or accuracy data sets is forbidden.

\subsection{Load Generator}
\label{sec:benchmarks:loadgen}

\begin{figure}[t!]
    \centering
    \includegraphics[width=0.99\linewidth]{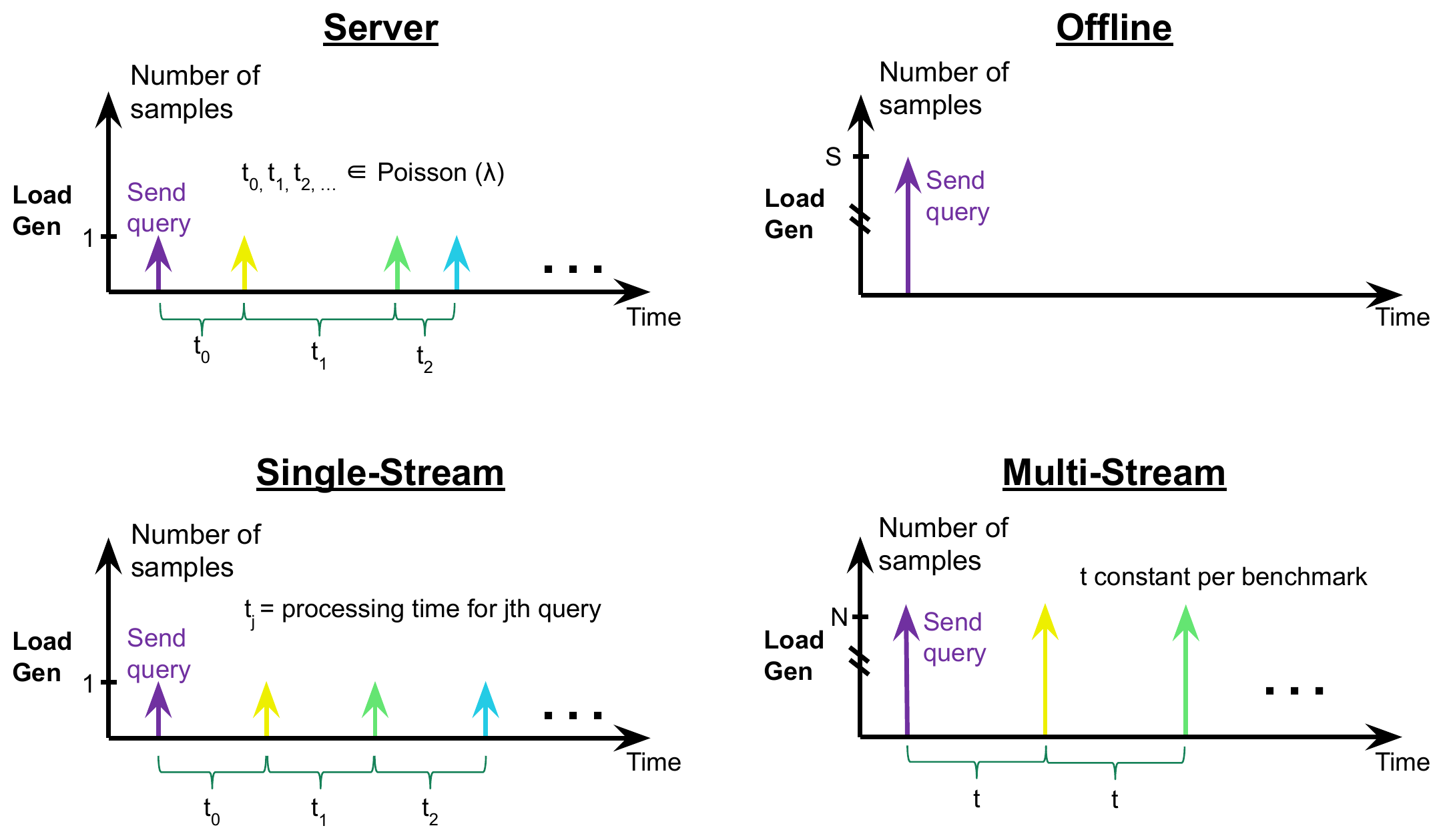}
    \vspace{-0.45cm}
    \caption{Timing and number of queries from the LoadGen.}
    \label{fig:Loadgen}
    \vskip -0.15in
\end{figure}

The LoadGen is a traffic generator for MLPerf Inference that loads the SUT and measures performance. Its behavior is controlled by a configuration file it reads at the start of the run. The LoadGen produces the query traffic according to the rules of the previously described scenarios (i.e., single-stream, multistream, server, and offline). Additionally, it collects information for logging, debugging, and postprocessing the data. It records queries and responses from the SUT, and at the end of the run, it reports statistics, summarizes the results, and determines whether the run was valid. Figure~\ref{fig:Loadgen} shows how the LoadGen creates query traffic for each scenario. In the server scenario, for instance, it issues queries in accordance with a Poisson distribution to mimic a server's query-arrival rates. In the single-stream case, it issues a query to the SUT and waits for completion of that query before issuing another. 

At startup, the LoadGen requests that the SUT load data-set samples into memory. The SUT may load them into DRAM as an untimed operation and perform other timed operations as the rules stipulate. These untimed operations include but are not limited to compilation, cache warmup, and preprocessing. The SUT signals the LoadGen when it is ready to receive the first query; a query is a request for inference on one or more samples. The LoadGen sends queries to the SUT in accordance with the selected scenario. Depending on that scenario, it can submit queries one at a time, at regular intervals, or in a Poisson distribution. The SUT runs inference on each query and sends the response back to the LoadGen, which either logs the response or discards it. After the run, an accuracy script checks the logged responses to determine whether the model accuracy is within tolerance.

The LoadGen has two primary operating modes: {accuracy} and {performance}. Both are necessary to validate MLPerf submissions. In accuracy mode, the LoadGen goes through the entire data set for the ML task. The model's task is to run inference on the complete data set. Afterward, accuracy results appear in the log files, ensuring the model met the required quality target. In performance mode, the LoadGen avoids going through the entire data set, as the system's performance can be determined by subjecting it to enough data-set samples.

We designed the LoadGen to flexibly handle changes to the benchmark suite. MLPerf Inference has an interface between the SUT and LoadGen so it can handle new scenarios and experiments in the LoadGen and roll them out to all models and SUTs without extra effort. Doing so also facilitates compliance and auditing, since many technical rules about query arrivals, timing, and accuracy are implemented outside of submitter code. We achieved this feat by decoupling the LoadGen from the benchmarks and the internal representations (e.g., the model,  scenarios, and quality and latency metrics). The LoadGen implementation is a standalone C++ module.

The decoupling allows the LoadGen to support various language bindings, permitting benchmark implementations in any language. The LoadGen supports Python, C, and C++ bindings; additional bindings can be added. Another benefit of decoupling the LoadGen from the benchmark is that the LoadGen is extensible to support more scenarios, such as a multitenancy mode where the SUT must continuously serve multiple models while maintaining QoS constraints.

Moreover, placing the performance-measurement code outside of submitter code fits with MLPerf's goal of end-to-end system benchmarking. The LoadGen therefore measures the holistic performance of the entire SUT rather than any individual part. Finally, this condition enhances the benchmark's realism: inference engines typically serve as black-box components of larger systems.

\subsection{Data Set}

We employ standard and publicly available data sets to ensure the community can participate. 
We do not host them directly, however. Instead, MLPerf downloads the data set before LoadGen uses it to run the benchmark. Table~\ref{tab:ml_task} lists the data sets that we selected for each of the benchmarks.

\subsection{Accuracy Checker}
\label{sec:loadgen_accuracy}

The LoadGen also has features that ensure the submission system complies with the rules. In addition, it can self-check to determine whether its source code has been modified during the submission process. To facilitate validation, the submitter provides an experimental config file that allows use of non-default LoadGen features. Details are in Section~\ref{sec:review_and_validation}.

\section{Submission-System Evaluation}
\label{sec:submission}

In this section, we describe the submission, review, and reporting process. Participants can submit results to different divisions and categories. All submissions are peer reviewed for validity. Finally, we describe how the results are reported.

\subsection{Result Submissions, Divisions, and Categories}\label{sec:sub}

A result submission contains information about the SUT: performance scores, benchmark code, a system-description file that highlights the SUT's main configuration characteristics (e.g., accelerator count, CPU count, software release, and memory system), and LoadGen log files detailing the performance and accuracy runs for a set of task and scenario combinations. All this data is uploaded to a public GitHub repository for peer review and validation before release.

MLPerf Inference is a suite of tasks and scenarios that ensures broad coverage, but a submission can contain a subset of them. Many traditional benchmarks, such as SPEC~CPU, require submissions for all components. This approach is logical for a general-purpose processor that runs arbitrary code, but ML systems are often highly specialized. 


\textbf{Divisions.} MLPerf Inference has two divisions for submitting results: closed and open. Participants can send results to either or both, but they must use the same data set.

The closed division enables comparison of different systems. Submitters employ the same models, data sets, and quality targets to ensure comparability across wildly different architectures. This division requires preprocessing, postprocessing, and a model that is equivalent to the reference implementation. It also permits calibration for quantization (using the calibration data set we provide) and prohibits retraining.

The open division fosters innovation in ML systems, algorithms, optimization, and hardware/software co-design. Perticipants must still perform the same ML task, but they may change the model architecture and the quality targets. This division allows arbitrary pre- and postprocessing and arbitrary models, including techniques such as retraining. In general, submissions are directly comparable neither with each other nor with closed submissions. Each open submission must include documentation about how it deviates from the closed division.

\textbf{Categories.}
Following MLPerf Training, participants classify their submissions into one of three categories on the basis of hardware and software availability: available; preview; and research, development, or other systems (RDO). This categorization helps consumers identify the systems' maturity and whether they are readily available (for rent or purchase).

\subsection{Result Review}
\label{sec:review_and_validation}

A challenge of benchmarking inference systems is that many include proprietary and closed-source components, such as inference engines and quantization flows, that make peer review difficult. To accommodate these systems while ensuring reproducible results that are free from common errors, we developed a validation suite to assist with peer review. These validation tools perform experiments that help determine whether a submission complies with the defined rules.




\textbf{Accuracy verification.} The purpose of this test is to ensure valid inferences in performance mode. By default, the results that the inference system returns to the LoadGen are not logged and thus are not checked for accuracy. This choice reduces or eliminates processing overhead to allow accurate measurement of the inference system's performance. In this test, results returned from the SUT to the LoadGen are logged randomly. The log is checked against the log generated in accuracy mode to ensure consistency.

\textbf{On-the-fly caching detection.} The LoadGen produces queries by randomly selecting query samples with replacement from the data set, and inference systems may receive queries with duplicate samples. This duplication is likely for high-performance systems that process many samples relative to the data-set size. To represent realistic deployments, the rules prohibit caching of queries and intermediate data. The test has two parts: the first generates queries with unique sample indices, and the second generates queries with duplicate sample indices. It measures performance in each case. The way to detect caching is to determine whether the test with duplicate sample indices runs significantly faster than the test with unique sample indices.

\textbf{Alternate-random-seed testing.} Ordinarily, the LoadGen produces queries on the basis of a fixed random seed. Optimizations based on that seed are prohibited. The alternate-random-seed test replaces the official random seed with alternates and measures the resulting performance.

\textbf{Custom data sets.} In addition to the LoadGen's validation features, we use custom data sets to detect result caching. MLPerf Inference validates this behavior by replacing the reference data set with a custom data set. We measure the quality and performance of the system operating on the latter and compare the results with operation on the former.

\subsection{Result Reporting}\label{sec:reporting}

MLPerf Inference provides no ``summary score.'' Benchmarking efforts often elicit a strong desire to distill the capabilities of a complex system to a single number and thereby enable comparison of different systems. But not all ML tasks are equally important for all systems, and the job of weighting some more heavily than others is highly subjective. At best, weighting and summarization are driven by the submitter catering to customer needs, as some systems may be designed for specific ML tasks. For instance, a system may be highly optimized for vision rather than for translation. In such cases, averaging the results across all tasks makes no sense, as the submitter may not be targeting those markets.

\section{Benchmark Assessment}
\label{sec:results}

On October 11, we put the inference benchmark to the test. We received from 14 organizations more than 600 submissions in all three categories (available, preview, and RDO) across the closed and open divisions. The results are the most extensive corpus of inference performance data available to the public, covering a range of ML tasks and scenarios, hardware architectures, and software run times. Each has gone through extensive review before receiving approval as a valid MLPerf result. After review, we cleared 595 results as valid. In this section, we assess the closed-division results on the basis of our inference-benchmark objectives:


\begin{itemize}

    \item Pick representative workloads for reproducibility, and allow everyone to access them (Section \ref{sec:results_task_coverage}).

    \item Identify usage scenarios for realistic evaluation (Section~\ref{sec:results_scenarios}).
    
    \item Set permissive rules that allow submitters to showcase both hardware and software capabilities (Sections~\ref{sec:results_diversity_arch} and \ref{sec:results_diversity_systems}).
   
    \item Describe a method that allows the benchmarks to change (Section~\ref{sec:results_open_division}).
\end{itemize}

\subsection{Task Coverage}
\label{sec:results_task_coverage}

\begin{figure}[t]
\centering
    {\includegraphics[width=\columnwidth]{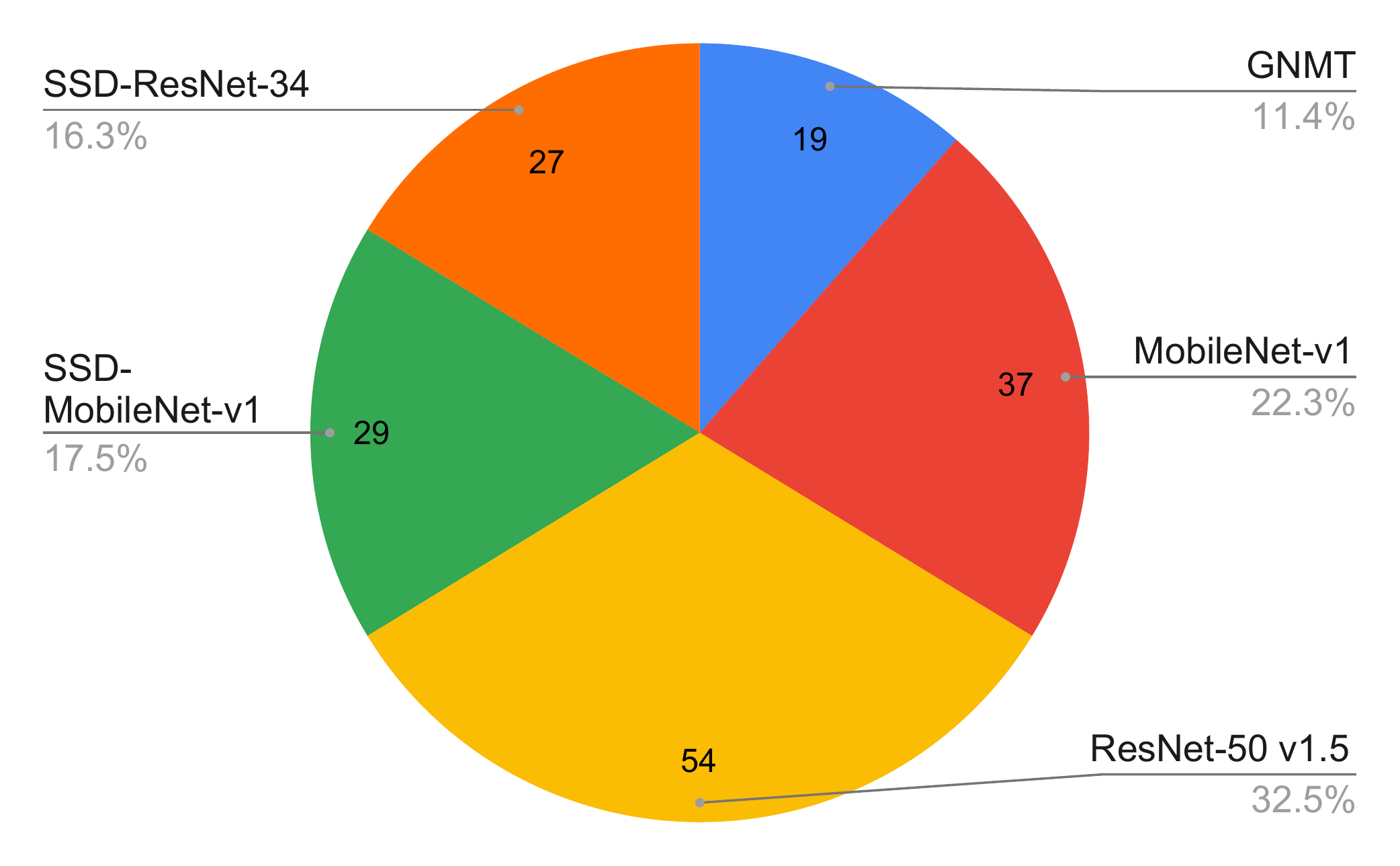}}
    \vspace{-0.45cm}
    \caption{Results from the closed division. The distribution indicates we selected representative workloads for the benchmark's initial release.}
    \label{fig:closed_division}
\end{figure}

Because we allow submitters to pick any task to evaluate their system's performance, the distribution of results across tasks can reveal whether those tasks are of interest to ML-system vendors. We analyzed the submissions to determine the overall coverage. Figure~\ref{fig:closed_division} shows the breakdown for the tasks and models in the closed division. Although the most popular model was---unsurprisingly---ResNet-50 v1.5, it was just under three times as popular as GNMT, the least popular model. This small spread and the otherwise uniform distribution suggests we selected a representative set of tasks.

In addition to selecting representative tasks, another goal is to provide vendors with varying quality and performance targets. The ideal ML model may differ with the use case (as Figure~\ref{fig:ml_model} shows, a vast range of models can target a given task). Our results reveal that vendors equally supported different models for the same task because each model has unique quality and performance tradeoffs. In the case of object detection, we saw about the same number of submissions for both SSD-MobileNet-v1 and SSD-ResNet-34.


\begin{table}[b!] 
 \vskip -0.05in
 \caption{High coverage of models and scenarios.}
 \vspace{-0.45cm}
 \label{tab:submissions}
 \vskip 0.05in
  \begin{center}
 \begin{small}
 \begin{sc}
 \begin{adjustbox}{width=\columnwidth, center}
\begin{tabular}{c | c | c |c | c }
\toprule
 & \textbf{Single-Stream} & \textbf{Multistream} & \textbf{Server} & \textbf{Offline} \\
\midrule
\textbf{GNMT} & 2 & 0 & 6 & 11  \\
\hline
\textbf{MobileNet-v1} & 18 & 3 & 5 & 11  \\
\hline
\textbf{ResNet-50 v1.5} & 19 & 5 & 10 & 20 \\
\hline
\textbf{SSD-MobileNet-v1} & 8 & 3 & 5 & 13 \\
\hline
\textbf{SSD-ResNet-34} & 4 & 4 & 7 & 12 \\
\hline\hline
\textbf{Total} & \textbf{51} & \textbf{15} & \textbf{33} & \textbf{67} \\

\bottomrule
\end{tabular}
\end{adjustbox}
\end{sc}
\end{small}
\end{center}
\vskip -0.15in
\end{table}

\subsection{Scenario Usage}
\label{sec:results_scenarios}

We aim to evaluate systems in realistic use cases---a major motivator for the LoadGen and scenarios. To this end, Table~\ref{tab:submissions} shows the distribution of results among the various task and scenario combinations. Across all tasks, the single-stream and offline scenarios are the most widely used and are also the easiest to optimize and run. Server and multistream were more complicated and had longer run times because of the QoS requirements and more-numerous queries. GNMT garnered no multistream submissions, possibly because the constant arrival interval is unrealistic in machine translation. Therefore, it was the only model and scenario combination with no submissions.

\begin{figure}[t!]
\centering
    {\includegraphics[width=.85\columnwidth]{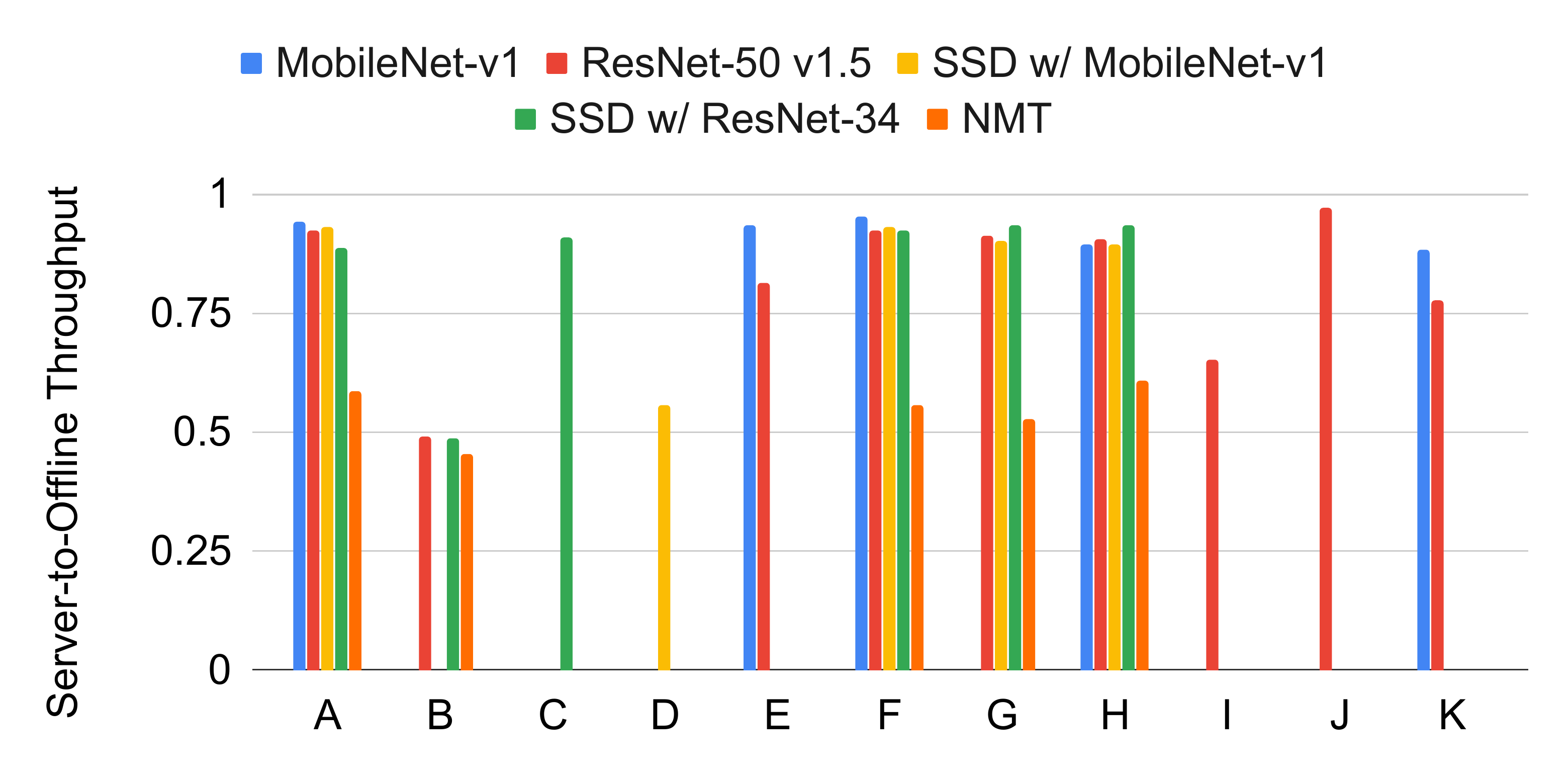}}
    \caption{Throughput degradation from server scenario (which has a latency constraint) for 11 arbitrarily chosen systems from the closed division. The server-scenario performance for each model is normalized to the performance of that same model in the offline scenario. A score of 1 corresponds to a model delivering the same throughput for the offline and server scenarios. Some systems, including C, D, I, and J, lack results for certain models because participants need not submit results for all models.}
    \label{fig:scenarios}
    \vspace{-10pt}
\end{figure}

The realistic MLPerf Inference scenarios are novel and illustrate many important and complex performance considerations that architects face but that studies often overlook. Figure~\ref{fig:scenarios} demonstrates that all systems deliver less throughput for the server scenario than for the offline scenario owing to the latency constraint and attendant suboptimal batching. Optimizing for latency is challenging and underappreciated.

Not all systems handle latency constraints equally well, however. For example, system B loses about 50\% or more of its throughput for all three models, while system A loses as much as 40\% for NMT, but approximately 10\% for the vision models. The throughput-degradation differences may be a result of a hardware architecture optimized for low batch size or more-effective dynamic batching in the inference engine and software stack---or, more likely, a combination of the two. Even with this limited data, one clear implication is that a performance comparison with unconstrained latency has little bearing on a latency-constrained scenario. Therefore, performance analysis should ideally include both.

Additionally, the performance impact of latency constraints varies with network type. Across all five systems with NMT results, the throughput reduction for the server scenario is 39--55\%. In contrast, the throughput reduction for ResNet-50 v1.5 varies from 3\% to 35\% with an average of about 20\%, and the average for MobileNet-v1 is under 10\%. The vast throughput-reduction differences likely reflect some combination of NMT's variable text input, more-significant software-stack optimization, and NMT's more-complex network architecture. A second lesson from this data is that the impact of latency constraints on different models extrapolates poorly. Even among classification models, the average performance loss for ResNet-50 v1.5 is approximately double that of MobileNet-v1.

\subsection{Processor Types and Software Frameworks}
\label{sec:results_diversity_arch}

\begin{figure}[t!]
\centering
{\includegraphics[width=.85\columnwidth]{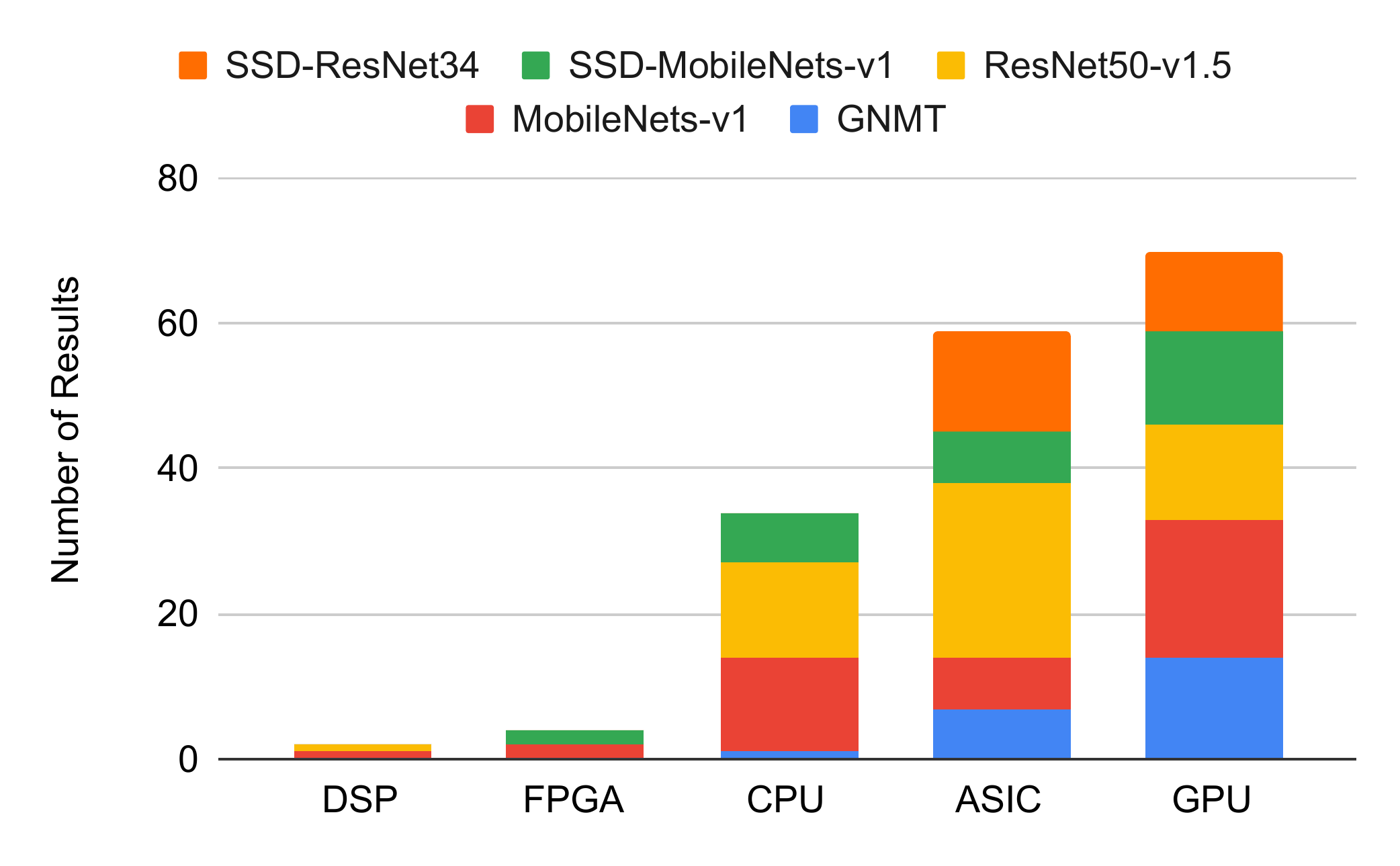}}
\vspace{-0.45cm}
    \caption{Results from the closed division. They cover almost every kind of processor architecture---CPUs, GPUs, DSPs, FPGAs, and ASICs.}
    \vspace{-0.25cm}
    \label{fig:submitted}
\end{figure}

A variety of platforms can employ ML solutions, from fully general-purpose CPUs to programmable GPUs and DSPs, FPGAs, and fixed-function accelerators. Our results reflect this diversity. Figure~\ref{fig:submitted} shows that the MLPerf Inference submissions covered most hardware categories, indicating our v0.5 benchmark suite and method can evaluate any processor architecture.

Many ML software frameworks accompany the various processor types. Table~\ref{tab:frameworks} shows the frameworks for benchmarking the hardware platforms. ML software plays a vital role in unleashing the hardware's performance. Some run times are designed to work with certain hardware types to fully harness their capabilities; employing the hardware without the corresponding framework may still succeed, but the performance may fall short of the hardware's potential. The table shows that CPUs have the most framework diversity and that TensorFlow has the most architectural variety.

\begin{table}[t!]
 \caption{Framework versus hardware architecture.}
 \vspace{-0.45cm}
 \label{tab:frameworks}
  \begin{center}
 \begin{sc}
\begin{adjustbox}{width=\columnwidth, center}
\fontsize{3}{4.0}
\selectfont
\begin{tabular}{c | c | c | c | c | c }
\toprule
 & \textbf{ASIC} & \textbf{CPU} & \textbf{DSP}  & \textbf{FPGA} & \textbf{GPU} \\
\midrule
\textbf{Arm NN} &  & X &  &  & X \\
\hline
\textbf{FuriosaAI} & & & & X & \\
\hline
\textbf{Hailo SDK} & X & & & &  \\
\hline
\textbf{HanGuang AI} & X & & & &  \\
\hline
\textbf{ONNX} &  & X & & & \\
\hline
\textbf{OpenVino} &  & X & & & \\
\hline
\textbf{PyTorch} &  & X & & & \\
\hline
\textbf{SNPE} & & & X & &  \\
\hline
\textbf{Synapse} & X &  & & &  \\
\hline
\textbf{TensorFlow} & X & X & & & X \\
\hline
\textbf{TensorFlow Lite} & & X & & & \\
\hline
\textbf{TensorRT} & & & & & X \\

\bottomrule
\end{tabular}
\end{adjustbox}
\end{sc}
\end{center}
\vskip -0.15in
\end{table}

\subsection{System Diversity}
\label{sec:results_diversity_systems}

The submissions cover a broad power and performance range, from mobile and edge devices to cloud computing. The performance delta between the smallest and largest inference systems is four orders of magnitude, or about 10,000$\times$. 

\begin{figure}[t]
\centering
{\includegraphics[trim=65 50 70 70, clip, width=\linewidth]{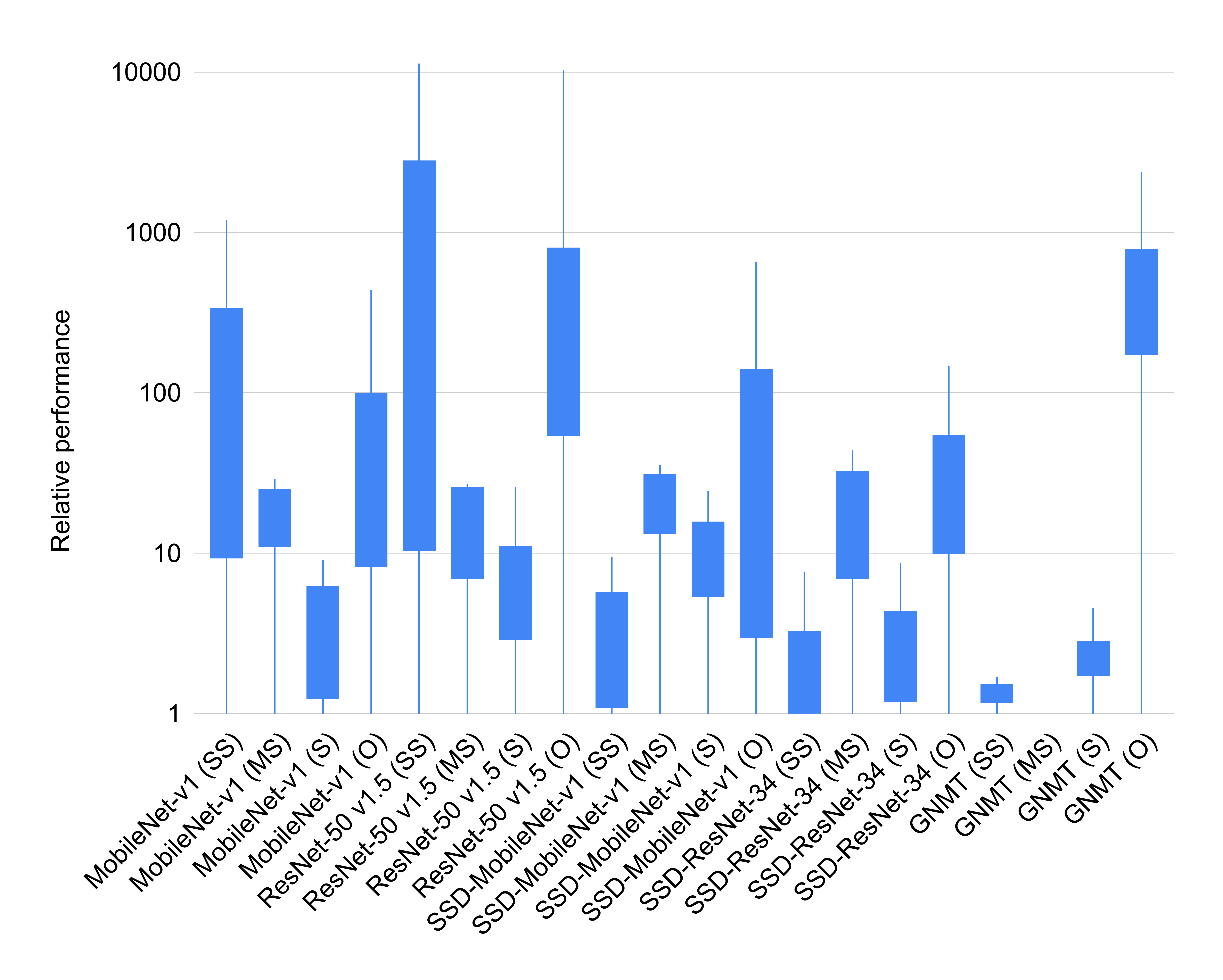}}
\vspace{-0.45cm}
    \caption{Performance for different models in the single-stream (SS), multistream (MS), server (S), and offline (O) scenarios. Scores are relative to the performance of the slowest system for the particular scenario.}
    \vspace{-0.25cm}
    \label{fig:violin}
\end{figure}

Figure~\ref{fig:violin} shows the results across all tasks and scenarios except for GNMT~(MS), which had no submissions. In cases such as the MobileNet-v1 single-stream scenario~(SS), ResNet-50 v1.5~(SS), and SSD-MobileNet-v1 offline~(O), systems exhibit a large performance difference (100$\times$). Because these models have many applications, the systems that target them cover everything from low-power embedded devices to high-performance servers. GNMT server (S) exhibits much less performance variation among systems.




The broad performance range implies that the tasks we initially selected for MLPerf Inference v0.5 are general enough to represent many use cases and market segments. The wide array of systems also indicates that our method (the LoadGen, metrics, etc.) is widely applicable.

\subsection{Open Division}
\label{sec:results_open_division}
We received 429 results in the less restrictive open division. A few highlights include 4-bit quantization to boost performance, exploration of various models (instead of the reference model) to perform the task, and high throughput under latency bounds tighter than what the closed-division rules stipulate.

We also saw submissions that pushed the limits of mobile-chipset performance. Typically, vendors use one accelerator at a time. We are seeing instances of multiple accelerators working concurrently to deliver high throughput in a multistream scenario---a rarity in conventional mobile situations. Together these results show the open division is encouraging the industry to push system limits.

\section{Lessons Learned}
\label{sec:conclusion}

Over the course of a year we have learned several lessons and identified opportunities for improvement, which we present here.

\subsection{Models: Breadth vs. Use-Case Depth}
\label{sec:lessons_breadth}


 
Balancing the breadth of applications (e.g., image recognition, objection detection, and translation) and models (e.g., CNNs and LSTMs) and the depth of the use cases (Table~\ref{tab:scenarios}) is important to industry. We therefore implemented 4 versions of each benchmark, 20 in total. Limited resources and the need for speedy innovation prevented us from including more applications (e.g., speech recognition and recommendation) and models (e.g., Transformers~\cite{vaswani2017attention}, BERT~\cite{devlin2018bert}, and DLRM~\cite{dlrm-training,dlrm-inference}), but we aim to add them soon.

\subsection{Metrics: Latency vs. Throughput}

Latency and throughput are intimately related, and considering them together is crucial: we use latency-bounded throughput (Table~\ref{tab:latency}). A system can deliver excellent throughput yet perform poorly when latency constraints arise. For instance, the difference between the offline and server scenarios is that the latter imposes a latency constraint and implements a nonuniform arrival rate. The result is lower throughput because large input batches become more difficult to form. For some systems, the server scenario's latency constraint reduces performance by as little as 3\% (relative to offline); for others, the loss is much greater (50\%).


\subsection{Data Sets: Public vs. Private}
The industry needs larger and better-quality public data sets for ML-system benchmarking. After surveying various industry and academic scenarios, we found for the SSD-large use case a wide spectrum of input-image sizes, ranging roughly from 200x200 to 8 MP. We settled on two resolutions as use-case proxies: small images, where 0.09 MP (300x300) represents mobile and some data-center applications, and large images, where 1.44 MP (1,200x1,200) represents robotics (including autonomous vehicles) and high-end video analytics. In practice, however, SSD-large employs an upscaled (1,200x1,200) COCO data set (Table~\ref{tab:ml_task}), as dictated by the lack of good public detection data sets with large images. Some such data sets exist, but they are less than ideal. ApolloScape~\cite{wang2019apolloscape}, for example, contains large images, but it lacks bounding-box annotations and its segmentation annotations omit labels for some object pixels (e.g., when an object has pixels in two or more noncontiguous regions, owing to occlusion). Generating the bounding-box annotations is therefore difficult in these cases. The Berkeley DeepDrive~\cite{yu2018bdd100k} images are lower in resolution (720p). We need more data-set generators to address these issues.

\subsection{Performance: Modeled vs. Measured}
Although it is common practice, characterizing a network's computational difficulty on the basis of parameter size or operator count (Figure~\ref{fig:ml_model}) can be an oversimplification. For example, 10 systems in the offline and server scenarios computed the performance of both SSD-ResNet-34 and SSD-MobileNet-v1. The former requires 175$\times$ more operations per image, but the actual throughput is only 50--60$\times$ less. This consistent 3$\times$ difference between the operation count and the observed performance shows how network structure can affect performance.


\subsection{Process: Audits and Auditability}



Because submitters can reimplement the reference benchmark to maximize their system's capabilities, the result-review process (Section~\ref{sec:review_and_validation}) was crucial to ensuring validity and reproducibility. We found about 40 issues in the approximately 180 results from the closed division. We ultimately released only 166 of these results. Issues ranged from failing to meet the quality targets (Table~\ref{tab:ml_task}), latency bounds (Table~\ref{tab:latency}), and query requirements (Table~\ref{tab:task_queries}) to inaccurately interpreting the rules. Thanks to the LoadGen's accuracy checkers (Section~\ref{sec:loadgen_accuracy}) and submission-checker scripts, we identified many issues automatically. The checkers also reduced the burden so only about three engineers had to comb through the submissions. In summary, since the diversity of options at every level of the ML inference stack is complicated (Figure~\ref{fig:hw_sw_stack}), we  found auditing and auditability to be necessary for ensuring result integrity and reproducibility.

\section{Prior Art in AI/ML Benchmarking}
\label{sec:priorart}

MLPerf strives to incorporate and build on the best aspects of prior work while also including community input. 


\textbf{AI Benchmark.} AI Benchmark~\cite{ignatov2019ai} is arguably the first mobile-inference benchmark suite. Its results and leaderboard focus on Android smartphones and only measure latency. The suite provides a summary score, but it fails to explicitly specify quality targets. We aim at a variety of devices (our submissions range from IoT devices to smartphones and edge/server-scale systems) as well as four scenarios per benchmark.


\textbf{EEMBC MLMark.} EEMBC MLMark~\cite{eembc} measures the performance and accuracy of embedded inference devices. It also includes image-classification and object-detection tasks, as MLPerf does, but it lacks use-case scenarios. MLMark measures performance at explicit batch sizes, whereas MLPerf allows submitters to choose the best batch sizes for different scenarios. Also, the former imposes no target-quality restrictions, whereas the latter does impose restrictions.


\textbf{Fathom.} Fathom~\cite{adolf2016fathom} provides a suite of models that incorporate several layer types (e.g., convolution, fully connected, and RNN). Still, it focuses on throughput rather than accuracy. Like Fathom, we include a suite of models that comprise various layers. Compared with Fathom, MLPerf provides both PyTorch and TensorFlow reference implementations for optimization, and it  introduces a variety of inference scenarios with different performance metrics.


\textbf{AI Matrix.} AI Matrix~\cite{aimatrix} is Alibaba's AI-accelerator benchmark. It uses microbenchmarks to cover basic operators such as matrix multiplication and convolution, it measures performance for fully connected and other common layers, it includes full models that closely track internal applications, and it offers a synthetic benchmark to match the characteristics of real workloads. MLPerf has a smaller model collection and focuses on simulating scenarios using the LoadGen.

\textbf{DeepBench.} Microbenchmarks such as DeepBench~\cite{baidu2017deepbench} measure the library implementation of kernel-level operations (e.g., 5,124$\times$700$\times$2,048 GEMM) that are important for performance in production models. They are useful for efficient development but fail to address the complexity of full models.

\textbf{TBD (Training Benchmarks for DNNs).} TBD~\cite{zhu2018tbd} is a joint project of the University of Toronto and Microsoft Research that focuses on ML training. It provides a wide spectrum of ML models in three frameworks (TensorFlow, MXNet, and CNTK), along with a powerful tool chain for their improvement. It focuses on evaluating GPU performance and only has one full model (Deep Speech 2) that covers inference.

\textbf{DAWNBench.} DAWNBench~\cite{coleman2017dawnbench} was the first multi-entrant benchmark competition to measure the end-to-end performance of deep-learning systems. It allowed optimizations across model architectures, optimization procedures, software frameworks, and hardware platforms. DAWNBench inspired MLPerf, but we offer more tasks, models, and scenarios.


\section{Conclusion}
\label{sec:conclusion}

MLPerf Inference’s core contribution is a comprehensive framework for measuring ML inference performance across a spectrum of use cases. We briefly summarize the three main aspects of inference benchmarking here.

\textbf{Performance metrics.} To make fair, apples-to-apples comparisons of AI systems, consensus on performance metrics is critical. We crafted a collection of such metrics: latency, latency-bounded throughput, throughput, and maximum number of inferences per query---all subject to a predefined accuracy target and some likelihood of achieving that target.

Latency or inference-execution time is often the metric that system and architecture designers employ. Instead, we identify latency-bounded throughput as a measure of inference performance in industrial use cases, representing data-center inference processing. Although this metric is common for data-center CPUs, we introduce it for data-center ML accelerators. Prior work often uses throughput or latency; the formulation in this paper reflects more-realistic deployment constraints.

\textbf{Accuracy/performance tradeoff.} ML systems often trade off between accuracy and performance. Prior art varies widely concerning acceptable inference-accuracy degradation. We consulted with domain experts from industry and academia to set both the accuracy and the tolerable degradation thresholds for MLPerf Inference, allowing distributed measurement and optimization of results to tune the accuracy/performance tradeoff. This approach standardizes AI-system design and evaluation, and it enables academic and industrial studies, which can now use the accuracy requirements of MLPerf Inference workloads to compare their efforts to industrial implementations and the established accuracy standards.

\textbf{Evaluation of AI inference accelerators.} An important contribution of this work is identifying and describing the metrics and inference scenarios (server, single-stream, multistream, and offline) in which AI inference accelerators are useful. An accelerator may stand out in one category while underperforming in another. Such a degradation owes to optimizations such as batching (or the lack thereof), which is use-case dependent. MLPerf introduces batching for three out of the four inference scenarios (server, multistream, and offline) across the five networks, and these scenarios can expose additional optimizations for AI-system development and research.

ML is still evolving. To keep pace with the changes, we established a process to maintain MLPerf~\cite{mattson2020mlperf}. We are updating the ML tasks and models for the next submission round while sticking to the established, fundamental ML-benchmarking method we describe in this paper. Nevertheless, we have much to learn. The MLPerf organization welcomes input and contributions; please visit \url{https://mlperf.org/get-involved}.




\bibliographystyle{IEEEtranS}
\bibliography{IEEEabrv,references}

\begin{thebibliography}{10}
\providecommand{\url}[1]{#1}
\csname url@samestyle\endcsname
\providecommand{\newblock}{\relax}
\providecommand{\bibinfo}[2]{#2}
\providecommand{\BIBentrySTDinterwordspacing}{\spaceskip=0pt\relax}
\providecommand{\BIBentryALTinterwordstretchfactor}{4}
\providecommand{\BIBentryALTinterwordspacing}{\spaceskip=\fontdimen2\font plus
\BIBentryALTinterwordstretchfactor\fontdimen3\font minus
  \fontdimen4\font\relax}
\providecommand{\BIBforeignlanguage}[2]{{%
\expandafter\ifx\csname l@#1\endcsname\relax
\typeout{** WARNING: IEEEtranS.bst: No hyphenation pattern has been}%
\typeout{** loaded for the language `#1'. Using the pattern for}%
\typeout{** the default language instead.}%
\else
\language=\csname l@#1\endcsname
\fi
#2}}
\providecommand{\BIBdecl}{\relax}
\BIBdecl

\bibitem{abadi2016tensorflow}
M.~Abadi, P.~Barham, J.~Chen, Z.~Chen, A.~Davis, J.~Dean, M.~Devin,
  S.~Ghemawat, G.~Irving, M.~Isard \emph{et~al.}, ``{TensorFlow: A System for
  Large-Scale Machine Learning},'' in \emph{OSDI}, vol.~16, 2016.

\bibitem{adolf2016fathom}
R.~Adolf, S.~Rama, B.~Reagen, G.-Y. Wei, and D.~Brooks, ``{Fathom: Reference
  Workloads for Modern Deep Learning Methods},'' in \emph{IEEE International
  Symposium on Workload Characterization (IISWC)}, 2016.

\bibitem{aimatrix}
{Alibaba}, ``Ai matrix.'' \url{https://aimatrix.ai/en-us/}, Alibaba, 2018.

\bibitem{amodei2018ai}
D.~Amodei and D.~Hernandez, ``Ai and compute,''
  \url{https://blog.openai.com/ai-and-compute/}, OpenAI, 2018.

\bibitem{apple2017coreml}
Apple, ``Core ml: Integrate machine learning models into your app,''
  \url{https://developer.apple.com/documentation/coreml}, Apple, 2017.

\bibitem{badrinarayanan2017segnet}
V.~Badrinarayanan, A.~Kendall, and R.~Cipolla, ``Segnet: A deep convolutional
  encoder-decoder architecture for image segmentation,'' \emph{IEEE
  transactions on pattern analysis and machine intelligence}, vol.~39, no.~12,
  2017.

\bibitem{bai2019onnx}
J.~Bai, F.~Lu, K.~Zhang \emph{et~al.}, ``Onnx: Open neural network exchange,''
  \url{https://github.com/onnx/onnx}, 2019.

\bibitem{baidu2017deepbench}
{Baidu}, ``{DeepBench: Benchmarking Deep Learning Operations on Different
  Hardware},'' \url{https://github.com/baidu-research/DeepBench}, 2017.

\bibitem{bianco2018benchmark}
S.~Bianco, R.~Cadene, L.~Celona, and P.~Napoletano, ``Benchmark analysis of
  representative deep neural network architectures,'' \emph{IEEE Access},
  vol.~6, 2018.

\bibitem{chen2015mxnet}
T.~Chen, M.~Li, Y.~Li, M.~Lin, N.~Wang, M.~Wang, T.~Xiao, B.~Xu, C.~Zhang, and
  Z.~Zhang, ``Mxnet: A flexible and efficient machine learning library for
  heterogeneous distributed systems,'' \emph{arXiv preprint arXiv:1512.01274},
  2015.

\bibitem{chetlur2018cudnn}
\BIBentryALTinterwordspacing
S.~Chetlur, C.~Woolley, P.~Vandermersch, J.~Cohen, J.~Tran, B.~Catanzaro, and
  E.~Shelhamer, ``cudnn: Efficient primitives for deep learning,'' \emph{CoRR},
  vol. abs/1410.0759, 2014. [Online]. Available:
  \url{http://arxiv.org/abs/1410.0759}
\BIBentrySTDinterwordspacing

\bibitem{chollet2015keras}
F.~Chollet \emph{et~al.}, ``Keras,'' \url{https://keras.io}, 2015.

\bibitem{chollet2017xception}
F.~Chollet, ``Xception: Deep learning with depthwise separable convolutions,''
  in \emph{Proceedings of Conference on Computer Vision and Pattern
  Recognition}, 2017.

\bibitem{coleman2017dawnbench}
C.~Coleman, D.~Narayanan, D.~Kang, T.~Zhao, J.~Zhang, L.~Nardi, P.~Bailis,
  K.~Olukotun, C.~R{\'e}, and M.~Zaharia, ``{DAWNBench: An End-to-End Deep
  Learning Benchmark and Competition},'' \emph{NeurIPS ML Systems Workshop},
  2017.

\bibitem{deng2009imagenet}
J.~Deng, W.~Dong, R.~Socher, L.-J. Li, K.~Li, and L.~Fei-Fei, ``Imagenet: A
  large-scale hierarchical image database,'' in \emph{2009 IEEE conference on
  computer vision and pattern recognition}.\hskip 1em plus 0.5em minus
  0.4em\relax Ieee, 2009.

\bibitem{devlin2018bert}
J.~Devlin, M.-W. Chang, K.~Lee, and K.~Toutanova, ``Bert: Pre-training of deep
  bidirectional transformers for language understanding,'' \emph{arXiv preprint
  arXiv:1810.04805}, 2018.

\bibitem{eembc}
{EEMBC}, ``Introducing the eembc mlmark benchmark,''
  \url{https://www.eembc.org/mlmark/index.php}, Embedded Microprocessor
  Benchmark Consortium, 2019.

\bibitem{goodfellow2014generative}
I.~Goodfellow, J.~Pouget-Abadie, M.~Mirza, B.~Xu, D.~Warde-Farley, S.~Ozair,
  A.~Courville, and Y.~Bengio, ``Generative adversarial nets,'' in
  \emph{Advances in Neural Information Processing Systems}, 2014.

\bibitem{dlrm-inference}
\BIBentryALTinterwordspacing
U.~Gupta, X.~Wang, M.~Naumov, C.~Wu, B.~Reagen, D.~Brooks, B.~Cottel, K.~M.
  Hazelwood, B.~Jia, H.~S. Lee, A.~Malevich, D.~Mudigere, M.~Smelyanskiy,
  L.~Xiong, and X.~Zhang, ``The architectural implications of facebook's
  dnn-based personalized recommendation,'' \emph{CoRR}, vol. abs/1906.03109,
  2019. [Online]. Available: \url{http://arxiv.org/abs/1906.03109}
\BIBentrySTDinterwordspacing

\bibitem{he2016deep}
K.~He, X.~Zhang, S.~Ren, and J.~Sun, ``Deep residual learning for image
  recognition,'' in \emph{Proceedings of Conference on Computer Vision and
  Pattern Recognition}, 2016.

\bibitem{howard2017mobilenets}
A.~G. Howard, M.~Zhu, B.~Chen, D.~Kalenichenko, W.~Wang, T.~Weyand,
  M.~Andreetto, and H.~Adam, ``Mobilenets: Efficient convolutional neural
  networks for mobile vision applications,'' \emph{arXiv preprint
  arXiv:1704.04861}, 2017.

\bibitem{hu2018squeeze}
J.~Hu, L.~Shen, and G.~Sun, ``Squeeze-and-excitation networks,'' in
  \emph{Proceedings of Computer Vision and Pattern Recognition}, 2018.

\bibitem{ignatov2019ai}
A.~Ignatov, R.~Timofte, A.~Kulik, S.~Yang, K.~Wang, F.~Baum, M.~Wu, L.~Xu, and
  L.~Van~Gool, ``Ai benchmark: All about deep learning on smartphones in
  2019,'' \emph{arXiv preprint arXiv:1910.06663}, 2019.

\bibitem{intel2018openvino}
Intel, ``Intel distribution of openvino toolkit,''
  \url{https://software.intel.com/en-us/openvino-toolkit}, Intel, 2018.

\bibitem{intel2018mkl}
Intel, ``Math kernel library,'' \url{https://software.intel.com/en-us/mkl},
  2018.

\bibitem{jia2014caffe}
Y.~Jia, E.~Shelhamer, J.~Donahue, S.~Karayev, J.~Long, R.~Girshick,
  S.~Guadarrama, and T.~Darrell, ``{Caffe: Convolutional Architecture for Fast
  Feature Embedding},'' in \emph{ACM International Conference on Multimedia},
  2014.

\bibitem{kanter2019supercomputing}
D.~Kanter, ``Supercomputing 19: Hpc meets machine learning,''
  \url{https://www.realworldtech.com/sc19-hpc-meets-machine-learning/}, real
  world technologies, 11 2019.

\bibitem{khudia2018fbgemm}
D.~S. Khudia, P.~Basu, and S.~Deng, ``Open-sourcing fbgemm for state-of-the-art
  server-side inference,''
  \url{https://engineering.fb.com/ml-applications/fbgemm/}, Facebook, 2018.

\bibitem{krizhevsky2012imagenet}
A.~Krizhevsky, I.~Sutskever, and G.~E. Hinton, ``Imagenet classification with
  deep convolutional neural networks,'' in \emph{Advances in Neural Information
  Processing Systems}, 2012.

\bibitem{lee2019device}
J.~Lee, N.~Chirkov, E.~Ignasheva, Y.~Pisarchyk, M.~Shieh, F.~Riccardi,
  R.~Sarokin, A.~Kulik, and M.~Grundmann, ``On-device neural net inference with
  mobile gpus,'' \emph{arXiv preprint arXiv:1907.01989}, 2019.

\bibitem{lee2019accelerating}
K.~Lee, V.~Rao, and W.~C. Arnold, ``Accelerating facebook’s infrastructure
  with application-specific hardware,''
  \url{https://engineering.fb.com/data-center-engineering/accelerating-infrastructure/},
  Facebook, 3 2019.

\bibitem{levine2018learning}
S.~Levine, P.~Pastor, A.~Krizhevsky, J.~Ibarz, and D.~Quillen, ``Learning
  hand-eye coordination for robotic grasping with deep learning and large-scale
  data collection,'' \emph{The International Journal of Robotics Research},
  vol.~37, no. 4-5, 2018.

\bibitem{lin2014microsoft}
T.-Y. Lin, M.~Maire, S.~Belongie, J.~Hays, P.~Perona, D.~Ramanan,
  P.~Doll{\'a}r, and C.~L. Zitnick, ``Microsoft coco: Common objects in
  context,'' in \emph{European conference on computer vision}.\hskip 1em plus
  0.5em minus 0.4em\relax Springer, 2014.

\bibitem{liu2016ssd}
W.~Liu, D.~Anguelov, D.~Erhan, C.~Szegedy, S.~Reed, C.-Y. Fu, and A.~C. Berg,
  ``Ssd: Single shot multibox detector,'' in \emph{European conference on
  computer vision}.\hskip 1em plus 0.5em minus 0.4em\relax Springer, 2016.

\bibitem{mattson2019mlperf}
P.~Mattson, C.~Cheng, C.~Coleman, G.~Diamos, P.~Micikevicius, D.~Patterson,
  H.~Tang, G.-Y. Wei, P.~Bailis, V.~Bittorf, D.~Brooks, D.~Chen, D.~Dutta,
  U.~Gupta, K.~Hazelwood, A.~Hock, X.~Huang, B.~Jia, D.~Kang, D.~Kanter,
  N.~Kumar, J.~Liao, D.~Narayanan, T.~Oguntebi, G.~Pekhimenko, L.~Pentecost,
  V.~J. Reddi, T.~Robie, T.~S. John, C.-J. Wu, L.~Xu, C.~Young, and M.~Zaharia,
  ``Mlperf training benchmark,'' \emph{arXiv preprint arXiv:1910.01500}, 2019.

\bibitem{mattson2020mlperf}
P.~Mattson, V.~J. Reddi, C.~Cheng, C.~Coleman, G.~Diamos, D.~Kanter,
  P.~Micikevicius, D.~Patterson, G.~Schmuelling, H.~Tang \emph{et~al.},
  ``Mlperf: An industry standard benchmark suite for machine learning
  performance,'' \emph{IEEE Micro}, vol.~40, no.~2, pp. 8--16, 2020.

\bibitem{mlperf2017resnet}
{MLPerf}, ``{ResNet in TensorFlow},''
  \url{https://github.com/mlperf/training/tree/master/image_classification/tensorflow/official},
  2019.

\bibitem{dlrm-training}
\BIBentryALTinterwordspacing
M.~Naumov, D.~Mudigere, H.~M. Shi, J.~Huang, N.~Sundaraman, J.~Park, X.~Wang,
  U.~Gupta, C.~Wu, A.~G. Azzolini, D.~Dzhulgakov, A.~Mallevich,
  I.~Cherniavskii, Y.~Lu, R.~Krishnamoorthi, A.~Yu, V.~Kondratenko, S.~Pereira,
  X.~Chen, W.~Chen, V.~Rao, B.~Jia, L.~Xiong, and M.~Smelyanskiy, ``Deep
  learning recommendation model for personalization and recommendation
  systems,'' \emph{CoRR}, vol. abs/1906.00091, 2019. [Online]. Available:
  \url{http://arxiv.org/abs/1906.00091}
\BIBentrySTDinterwordspacing

\bibitem{nvidiaYYYYtensorrt}
NVIDIA, ``Nvidia tensorrt: Programmable inference accelerator,''
  \url{https://developer.nvidia.com/tensorrt}, NVIDIA.

\bibitem{papineni2002bleu}
K.~Papineni, S.~Roukos, T.~Ward, and W.-J. Zhu, ``Bleu: a method for automatic
  evaluation of machine translation,'' in \emph{Proceedings of the 40th annual
  meeting on association for computational linguistics}.\hskip 1em plus 0.5em
  minus 0.4em\relax Association for Computational Linguistics, 2002.

\bibitem{paszke2017pytorch}
A.~Paszke, S.~Gross, S.~Chintala, G.~Chanan, E.~Yang, Z.~DeVito, Z.~Lin,
  A.~Desmaison, L.~Antiga, and A.~Lerer, ``Automatic differentiation in
  pytorch,'' 2017.

\bibitem{post2018call}
M.~Post, ``A call for clarity in reporting bleu scores,'' \emph{arXiv preprint
  arXiv:1804.08771}, 2018.

\bibitem{aixprt}
{Principled Technologies}, ``Aixprt community preview,''
  \url{https://www.principledtechnologies.com/benchmarkxprt/aixprt/}, 2019.

\bibitem{qualcommYYYYsnpe}
Qualcomm, ``Snapdragon neural processing engine sdk reference guide,''
  \url{https://developer.qualcomm.com/docs/snpe/overview.html}, Qualcomm.

\bibitem{sandler2018mobilenetv2}
M.~Sandler, A.~Howard, M.~Zhu, A.~Zhmoginov, and L.-C. Chen, ``Mobilenetv2:
  Inverted residuals and linear bottlenecks,'' in \emph{Proceedings of
  Conference on Computer Vision and Pattern Recognition}, 2018.

\bibitem{seide2016cntk}
F.~Seide and A.~Agarwal, ``Cntk: Microsoft's open-source deep-learning
  toolkit,'' in \emph{Proceedings of the 22nd ACM SIGKDD International
  Conference on Knowledge Discovery and Data Mining}.\hskip 1em plus 0.5em
  minus 0.4em\relax ACM, 2016.

\bibitem{tamhane2000statistics}
A.~Tamhane and D.~Dunlop, ``Statistics and data analysis: from elementary to
  intermediate,'' \emph{Prentice Hall}, 2000.

\bibitem{basicmi}
S.~Tang, ``Ai-chip,'' \url{https://basicmi.github.io/AI-Chip/}, 2019.

\bibitem{tokui2015chainer}
S.~Tokui, K.~Oono, S.~Hido, and J.~Clayton, ``Chainer: a next-generation open
  source framework for deep learning,'' in \emph{Proceedings of workshop on
  machine learning systems (LearningSys) in Neural Information Processing
  Systems (NeurIPS)}, vol.~5, 2015.

\bibitem{vaswani2017attention}
A.~Vaswani, N.~Shazeer, N.~Parmar, J.~Uszkoreit, L.~Jones, A.~N. Gomez,
  {\L}.~Kaiser, and I.~Polosukhin, ``Attention is all you need,'' in
  \emph{Advances in Neural Information Processing Systems}, 2017.

\bibitem{wang2019apolloscape}
P.~Wang, X.~Huang, X.~Cheng, D.~Zhou, Q.~Geng, and R.~Yang, ``The apolloscape
  open dataset for autonomous driving and its application,'' \emph{IEEE
  transactions on pattern analysis and machine intelligence}, 2019.

\bibitem{wmt2016}
\BIBentryALTinterwordspacing
WMT, ``First conference on machine translation,'' 2016. [Online]. Available:
  \url{http://www.statmt.org/wmt16/}
\BIBentrySTDinterwordspacing

\bibitem{wu2016google}
Y.~Wu, M.~Schuster, Z.~Chen, Q.~V. Le, M.~Norouzi, W.~Macherey, M.~Krikun,
  Y.~Cao, Q.~Gao, K.~Macherey \emph{et~al.}, ``Google's neural machine
  translation system: Bridging the gap between human and machine translation,''
  \emph{arXiv preprint arXiv:1609.08144}, 2016.

\bibitem{xie2017aggregated}
S.~Xie, R.~Girshick, P.~Doll{\'a}r, Z.~Tu, and K.~He, ``Aggregated residual
  transformations for deep neural networks,'' in \emph{Proceedings of
  Conference on Computer Vision and Pattern Recognition}, 2017.

\bibitem{xu2018pointfusion}
D.~Xu, D.~Anguelov, and A.~Jain, ``Pointfusion: Deep sensor fusion for 3d
  bounding box estimation,'' in \emph{Proceedings of Conference on Computer
  Vision and Pattern Recognition}, 2018.

\bibitem{yu2018bdd100k}
F.~Yu, W.~Xian, Y.~Chen, F.~Liu, M.~Liao, V.~Madhavan, and T.~Darrell,
  ``Bdd100k: A diverse driving video database with scalable annotation
  tooling,'' \emph{arXiv preprint arXiv:1805.04687}, 2018.

\bibitem{zhu2018tbd}
H.~Zhu, M.~Akrout, B.~Zheng, A.~Pelegris, A.~Jayarajan, A.~Phanishayee,
  B.~Schroeder, and G.~Pekhimenko, ``Benchmarking and analyzing deep neural
  network training,'' in \emph{IEEE International Symposium on Workload
  Characterization (IISWC)}, 2018.

\end{thebibliography}

\section*{Acknowledgements}
\label{sec:acks}

MLPerf Inference is the work of many individuals from multiple organizations. In this section, we acknowledge all those who helped produce the first set of results or supported the overall benchmark development.

\minihead{Alibaba T-Head:}
Zhi Cai,
 Danny Chen,
 Liang Han,
 Jimmy He,
 David Mao,
 Benjamin Shen,
 ZhongWei Yao,
 Kelly Yin,
 XiaoTao Zai,
 Xiaohui Zhao,
 Jesse Zhou,
 and
 Guocai Zhu.

\minihead{Baidu:}
Newsha Ardalani,
Ken Church,
and
Joel Hestness.

\minihead{Cadence:}
Debajyoti Pal.

\minihead{Centaur Technology:}
 Bryce Arden,
 Glenn Henry,
 CJ Holthaus,
 Kimble Houck,
 Kyle O'Brien,
 Parviz Palangpour,
 Benjamin Seroussi,
 and
 Tyler Walker.

\minihead{Dell EMC:}
Frank Han, Bhavesh Patel, Vilmara Rocio Sanchez,  and
Rengan Xu.

\minihead{dividiti:}
Grigori Fursin and Leo Gordon. 

\minihead{Facebook:}
Soumith Chintala, Kim Hazelwood, Bill Jia,  and
Sean Lee.

\minihead{FuriosaAI:}
Dongsun Kim and Sol Kim.

\minihead{Google:}
 Michael Banfield,
 Victor Bittorf,
 Bo Chen,
 Dehao Chen,
 Ke Chen,
 Chiachen Chou,
 Sajid Dalvi,
 Suyog Gupta,
 Blake Hechtman,
 Terry Heo,
 Andrew Howard,
 Sachin Joglekar,
 Allan Knies,
 Naveen Kumar,
 Cindy Liu,
 Thai Nguyen,
 Tayo Oguntebi,
 Yuechao Pan,
 Mangpo Phothilimthana, 
 Jue Wang,
 Shibo Wang,
 Tao Wang,
 Qiumin Xu,
 Cliff Young,
 Ce Zheng, 
 and
 Zongwei Zhou.
 
\minihead{Hailo:}
Ohad Agami, Mark Grobman,  and
Tamir Tapuhi.

\minihead{Intel:}
Md Faijul Amin,
 Thomas Atta-fosu,
 Haim Barad,
 Barak Battash,
 Amit Bleiweiss,
 Maor Busidan,
 Deepak R Canchi,
 Baishali Chaudhuri,
 Xi Chen,
 Elad Cohen,
 Xu Deng,
 Pradeep Dubey,
 Matthew Eckelman,
 Alex Fradkin,
 Daniel Franch,
 Srujana Gattupalli,
 Xiaogang Gu,
 Amit Gur,
 MingXiao Huang,
 Barak Hurwitz,
 Ramesh Jaladi,
 Rohit Kalidindi,
 Lior Kalman,
 Manasa Kankanala,
 Andrey Karpenko,
 Noam Korem,
 Evgeny Lazarev,
 Hongzhen Liu,
 Guokai Ma,
 Andrey Malyshev,
 Manu Prasad Manmanthan,
 Ekaterina Matrosova,
 Jerome Mitchell,
 Arijit Mukhopadhyay,
 Jitender Patil,
 Reuven Richman,
 Rachitha Prem Seelin,
 Maxim Shevtshov,
 Avi Shimalkovski,
 Dan Shirron,
 Hui Wu,
 Yong Wu,
 Ethan Xie,
 Cong Xu,
 Feng Yuan,
 and
 Eliran Zimmerman.

\minihead{MediaTek:}
Bing Yu.

\minihead{Microsoft:}
Scott McKay, Tracy Sharpe,  and
Changming Sun.

\minihead{Myrtle:}
Peter Baldwin.

\minihead{NVIDIA:}
Felix Abecassis,
 Vikram Anjur,
 Jeremy Appleyard,
 Julie Bernauer,
 Anandi Bharwani,
 Ritika Borkar,
 Lee Bushen,
 Charles Chen,
 Ethan Cheng,
 Melissa Collins,
 Niall Emmart,
 Michael Fertig,
 Prashant Gaikwad,
 Anirban Ghosh,
 Mitch Harwell,
 Po-Han Huang,
 Wenting Jiang,
 Patrick Judd,
 Prethvi Kashinkunti,
 Milind Kulkarni,
 Garvit Kulshreshta,
 Jonas Li,
 Allen Liu,
 Kai Ma,
 Alan Menezes,
 Maxim Milakov,
 Rick Napier,
 Brian Nguyen,
 Ryan Olson,
 Robert Overman,
 Jhalak Patel,
 Brian Pharris,
 Yujia Qi,
 Randall Radmer,
 Supriya Rao,
 Scott Ricketts,
 Nuno Santos,
 Madhumita Sridhara,
 Markus Tavenrath,
 Rishi Thakka,
 Ani Vaidya,
 KS Venkatraman,
 Jin Wang,
 Chris Wilkerson,
 Eric Work,
 and
 Bruce Zhan.

\minihead{Politecnico di Milano:}
Emanuele Vitali.

\minihead{Qualcomm:}
 Srinivasa Chaitanya Gopireddy,
 Pradeep Jilagam,
 Chirag Patel,
 Harris Teague,
 and
 Mike Tremaine.
 
\minihead{Samsung:}
Rama Harihara,
 Jungwook Hong,
 David Tannenbaum,
 Simon Waters,
 and
 Andy White.
 
\minihead{Stanford University:}
Peter Bailis and
Matei Zaharia.

\minihead{Supermicro:}
Srini Bala,
 Ravi Chintala,
 Alec Duroy,
 Raju Penumatcha,
 Gayatri Pichai,
 and
 Sivanagaraju Yarramaneni.
 
\minihead{Unaffiliated:}
Michael Gschwind and Justin Sang.

\minihead{University of California, Berkeley / Google:}
David Patterson.

\minihead{Xilinx:}
Ziheng Gao,
 Yiming Hu,
 Satya Keerthi Chand Kudupudi,
 Ji Lu,
 Lu Tian,
 and
 Treeman Zheng.

\end{document}